\documentclass[11pt]{article}

\usepackage[final]{acl}

\usepackage{times}
\usepackage{latexsym}
\usepackage{amsmath}
\usepackage{booktabs}
\usepackage{tikz}
\usepackage{pgfplots}
\usepackage{tabularx}
\usepackage{CJKutf8} %
\usepgfplotslibrary{groupplots}
\pgfplotsset{compat=1.18}

\usepackage[T1]{fontenc}

\usepackage[utf8]{inputenc}

\usepackage{microtype}

\usepackage{enumitem}

\usepackage{inconsolata}

\usepackage{graphicx}

\usepackage{amsmath,amssymb,amsthm}
\usepackage{booktabs}  %
\usepackage{graphicx}
\usepackage{hyperref}       %
\usepackage[capitalize,noabbrev]{cleveref}

\usepackage{algorithmic}
\usepackage{mathtools}
\usepackage{tcolorbox}
\usepackage{cleveref}
\usepackage{todonotes}

\newcommand{\R}{\mathbb{R}} %

\newcommand{\vertiii}[1]{{\left\vert\kern-0.25ex\left\vert\kern-0.25ex\left\vert #1 
    \right\vert\kern-0.25ex\right\vert\kern-0.25ex\right\vert}} %

\usepackage{bbding}
\usepackage{pifont}
\newcommand{\halftick}{\textcolor{brown}{\ding{51}}{\small\textcolor{brown}{\kern-0.7em\ding{55}}}}

\title{Studying quantization trade-offs for efficient inference deployment in machine translation}

\author{
 \textbf{Jim Zhao\textsuperscript{1,2}\thanks{Work done during internship at Aleph Alpha Research.}},
 \textbf{Sohir Maskey\textsuperscript{2}},
 \textbf{Koen Oostermeijer\textsuperscript{2}},
 \textbf{Douglas Orr\textsuperscript{3}},
 \textbf{Teryn Jones\textsuperscript{2}}
\\
\\
 \textsuperscript{1} University of Basel,
 \textsuperscript{2} Aleph Alpha Research,
 \textsuperscript{3} Graphcore,
\\
 \small{
   \textbf{Correspondence:} \href{mailto:jim.zhao@unibas.ch}{jim.zhao@unibas.ch}
 }
}

\begin{document}

\maketitle

\begin{abstract}
Deploying large language models in realistic server environments poses challenges, as the system needs to provide high-quality responses with low latency.
Quantization is a common approach to reduce the memory footprint and improve inference efficiency, yet its impact on latency and throughput is rarely evaluated under controlled, orchestration-level workloads.
In this work we study the quantization trade-offs of EuroLLM \citep{martins2025eurollm} across three model sizes ranging from 1.7B to 22B for efficient deployment on a single A100 or H100 GPU.
We demonstrate that combining a document-chunking strategy with W4A8 or W8A8 quantization improves the latency-throughput Pareto-curve under a wide range of workloads. 
Furthermore, since standard machine translation (MT) benchmarks rely on isolated sentences and fail to capture long-context dynamics, we introduce a document-level evaluation based on DocHPLT \cite{o2025dochplt} to assess how text chunking strategies affect translation quality under quantization.
Our results indicate that standard segment-level evaluation can potentially underestimate the interaction between quantization and long-context document translation, for some quantization formats, translation direction and models.
Overall, our experiments show that the trade-off between inference efficiency and translation quality depends not only on the quantization format, but also on the choice of text chunking strategy.

\end{abstract}

\section{Introduction}
    Large language models have revolutionized the field of machine learning and beyond, achieving close to or beyond human expert level across a wide range of domains, including coding \citep{coignion2024performance}, medicine \citep{liu2025generalist}, and machine translation \citep{zheng2026hy}. 
    Deploying these models efficiently remains challenging. Real-world serving systems must handle concurrent requests while balancing throughput and latency \citep{agrawal2024taming}. These constraints are particularly relevant when models are deployed on a limited number of GPUs. Quantization is a common approach to reduce model memory requirements and can substantially improve inference efficiency \citep{jarmusch2026microbenchmarking}.

    Most prior work studies the quality retention and inference benefits of quantization formats in isolation, primarily focusing on memory or token-level metrics. 
    In contrast, we analyze quantization trade-offs for MT inference under document-level serving constraints, focusing on end-to-end latency and throughput.
    We study these trade-offs on the EuroLLM {\citep{martins2025eurollm}} model family across three model sizes ranging from 1.7B to 22B parameters for deployment on a single A100 or H100 GPU, leading to the following contributions:

    \begin{itemize}[topsep=1pt, itemsep=1pt, partopsep=1pt, parsep=1pt]%
        \item We show via controlled closed-loop synthetic online benchmarks that combining W8A8 and W4A8 with a document-chunking strategy of 200-400 tokens provides the best efficiency gains for medium-sized documents across most of the considered deployment settings.
        \item Furthermore, we show that segment-level MT evaluation can potentially underestimate the quality degradation under quantization for certain translation directions and quantization formats.
        \item Concretely, for the Ukrainian $\to$ English translation direction we observe a stronger degradation of W4A16 with growing chunk size up to 1500 input prompt tokens for EuroLLM-1.7B and up to 4K input prompt tokens for EuroLLM-9B.  
    \end{itemize}

    The remaining paper is structured as follows: We review related work in \cref{sec:related_work}. In \cref{sec:Methods} we provide details about methods used in this work. In \cref{sec:experiments} we present our experimental results, which we further discuss in \cref{sec:discussion}. We discuss limitations in \cref{sec:limitations} and finally conclude in \cref{sec:conclusion}.

\section{Related work}\label{sec:related_work}

\subsection{Quantization Paradigms}
    Quantization can improve inference efficiency by reducing the precision of model weights, activations, or the KV cache. 
    Weight quantization reduces the model size and provides speed-ups under low batch sizes, where inference is often memory-bandwidth bound. 
    As batch size increases, arithmetic intensity grows and inference can become compute-bound, reducing the relative benefit of weight-only quantization alone.
    Combining weight quantization with activation quantization overcomes this limitation by leveraging hardware-native low-precision compute formats,
    such as INT8, FP8, MXFP4 \citep{rouhani2023microscaling}, NVFP4 \citep{xin2026quantization}, which enables arithmetic acceleration.
     Finally, KV quantization \citep{hooper2024kvquant} compresses the runtime memory footprint of the KV cache.
     This compression reduces memory bottlenecks under long sequences or large batch sizes, which might require CPU memory offloading.

    \paragraph{Post-training quantization (PTQ)}
    PTQ refers to methods, which quantize models after they have been fully trained. They typically use a small calibration dataset, although data-free approaches also exist \citep{nagel2019data}.
    Popular PTQ methods include GPTQ \citep{frantar2022gptq}, AWQ \citep{lin2025awq} or QQQ \citep{zhang2024qqq}.

\subsection{Efficient LLM Serving and Chunking}

There is a rich line of work focused on optimizing the deployment and serving of large language models. 
At the scheduling level, \citet{yu2022orca} introduced iteration-level scheduling, which is widely referred to as continuous batching, to efficiently process incoming requests of varying lengths. To address memory bottlenecks, \citet{kwon2023efficient} proposed PagedAttention to mitigate KV-cache fragmentation. 
These techniques are now standard in production-grade serving engines, such as vLLM \citep{kwon2023efficient} and vendor specific frameworks like TensorRT-LLM \citep{nvidia2023tensorrtllm}, which provide hardware-accelerated inference. 
Beyond engine-level memory management, inference efficiency can also be controlled at the input level through document chunking. While underexplored in MT, chunking strategies are heavily researched in Retrieval-Augmented Generation (RAG). For instance, \citet{qu2025semantic} challenged the benefits of semantic chunking over more efficient fixed-size chunking in RAGs.

\subsection{Document-level Machine Translation}

While widely used MT benchmarks, such as WMT24++ \citep{deutsch2025wmt24++} or Flores+ \citep{flores101-22} primarily evaluate translations at the segment level, document-level MT has a substantial research history. \citet{laubli-etal-2018-machine} demonstrate that evaluation protocols based on isolated sentences can lead to different conclusions from document-level evaluations when comparing professional and machine translations. 
\citet{lopes-etal-2020-document} compare several document-level context-aware neural MT approaches, and \citet{maruf2021survey} surveys methods and evaluation protocols for document-level MT. 
More recently, \citet{wang-etal-2023-document-level} evaluate GPT-3.5 \citep{ouyang2022training} and GPT-4 \citep{achiam2023gpt} for document-level translation and find that they
outperform commercial MT systems under human evaluation, highlighting the potential of decoder-only LLMs for this setting. Similarly, \citet{karpinska-iyyer-2023-large} find that translating complete literary paragraphs with GPT-3.5 \citep{ouyang2022training} yields higher-quality translations than translating each sentence independently across 18 language pairs. However, critical errors, including occasional content omissions, remain even when the model is given document-level context.

\subsection{Evaluation of Quantized LLMs}

\citet{li2024evaluating} evaluate quantized LLMs over a wide range of tasks, including long-context tasks, such as key-value retrieval and multi-document question answering. 
\citet{marchisio2024does} study how quantization affects multilingual LLMs, showing that, in line with our finding, automatic benchmarks often underestimate the quantization effects 
and that languages are disparately affected by quantization. 
Finally, in the context of MT, \citet{marie2025uneven} study how different PTQ methods affect translation quality across languages. They do not examine how quantization
interacts with document context or document-chunking strategies.\\\\
Our work connects these lines of research by jointly studying
quantization, document chunking, serving efficiency, and
document-level translation quality.

\section{Methods}\label{sec:Methods}

\subsection{Machine Translation Models}

We consider the EuroLLM \citep{martins2025eurollm} model family across three model sizes ranging from 1.7B to 22B. 
EuroLLM utilizes a standard dense decoder-only Transformer architecture.
Concretely, we use the instruction-tuned variants EuroLLM-1.7B-Instruct, EuroLLM-9B-Instruct-2512, and EuroLLM-22B-2512, where the first has a context length of 4096 and the latter two have a context length of 32K. All models use RoPE \citep{su2024roformer} with $\theta = 10.000$ for EuroLLM-1.7B-Instruct and $\theta = 1.000.000$ for EuroLLM-9B-Instruct-2512 and EuroLLM-22B-2512.
For brevity, we will refer to the EuroLLM models simply by their model size, e.g. EuroLLM-1.7B.

\subsection{Hardware and Format Constraints}

In this work, we consider three combinations of weight and activation quantization: W8A8 (8-bit weight and activation), W4A8 (4-bit weight and 8-bit activation), and W4A16 (4-bit weight-only). Further details on the concrete quantization scheme can be found in \cref{appendix:details_quantization}. While there is an ongoing debate regarding whether floating-point or integer formats should be preferred for LLMs \citep{zhang2024integer},
we observe that W8A8 often remains close to the translation quality of the unquantized BF16 baseline in most cases while benefiting from widely supported INT8 tensor cores. This allows the deployment of the quantized model across a broad range of GPUs (including the RTX4090, A100, and H100) using highly optimized compute kernels.\\
Furthermore, while more sophisticated 4-bit formats like MXFP4 \citep{rouhani2023microscaling} and NVFP4 \citep{abecassis2025pretraining} show promise, they are only natively supported on newer Blackwell architectures \citep{jarmusch2026microbenchmarking}. Finally, extreme low bit-widths (<4 bits) \citep{ma2024era, maskey20261, wang2023bitnet, wu2025bitnet} are excluded from this study, as they are primarily optimized for highly memory-constrained, on-device deployment rather than achieving maximum throughput across the wide range of concurrent batch sizes expected in a server setting.

\subsection{Post-Training Quantization}
We perform post-training quantization (PTQ) using GPTQ \citep{frantar2022gptq} and apply SmoothQuant \citep{xiao2023smoothquant} before performing GPTQ to reduce activation outliers before activation quantization for W8A8 and W4A8, where we choose a smooth strength of 0.8 for W8A8 and 0.4 for W4A8 quantization. For W4A16 GPTQ is applied directly. We performed GPTQ on a subset of 1024 samples subsampled from OPUS \citep{tiedemann2016opus}, where the target translations were generated by Seed-X \citep{cheng2025seed}.
All linear layers in the model, except for the embedding and head layer, are quantized using the same data format. 

\subsection{vLLM Engine}

We use vLLM \citep{kwon2023efficient} as our serving engine, which provides state-of-the-art kernel implementations, including W4A16, W4A8 and W8A8 kernels. 
Furthermore, it implements additional inference optimizations, such as PagedAttention \citep{kwon2023efficient} and Continuous Batching \citep{yu2022orca}. We can thus evaluate the benefit of quantized models for inference under near-production conditions, without inefficient kernel implementations potentially distorting the picture. 
We note that vLLM does not support W4A8 kernels on A100 by default, which is why we limit experiments using W4A8 on the H100 GPU.
We used vLLM version 0.26.0 for our experiments.

\subsection{Translation evaluation}

To study the translation quality retention, we benchmarked the EuroLLM-1.7B and EuroLLM-9B model on BF16, W4A16 and W8A8 on a subset of DocHPLT \citep{o2025dochplt} on a RTX4090 GPU.
We select EN $\leftrightarrow$ UK as an exemplary translation pair supported in EuroLLM and covered by DocHPLT, leading to two translation directions in total. 

\subsubsection{Segment-level translation}
For segment-level evaluation, we use chrF++ \citep{popovic2017chrf++} to compare the translation quality retained by the quantized model. Segments are translated independently, then translations and references are concatenated by document before computing chrF++, which scores translations in the range $[0,100]$, where a higher score is better.

\subsubsection{Document-level evaluation}

We construct document-level inputs from high-confidence aligned segments in DocHPLT.
DocHPLT \citep{o2025dochplt} contains parallel documents that are split into partially aligned source and reference alignment records.
To construct document-level inputs without breaking this alignment, we first filter documents by alignment records with Aligner-Score $\geq$ 0.5 and Bicleaner-Score $\geq$ 0.8, merge overlapping records into non-overlapping bilingual blocks, and preserve their original document order before chunking.
This ensures that the blocks in each documents are sufficiently well aligned. 
We then greedily concatenate complete blocks in the document order until the tokenized prompt reaches or exceeds a target prompt-length threshold $T$. As blocks are never split to avoid breaking existing alignment, the realized prompt length may exceed $T$. In addition, short trailing chunks are also retained when a document produces multiple chunks to obtain a full document translation.
We exclude a document at threshold $T$ only when the complete document forms a single prompt shorter than $0.8T$, in order to prevent short documents from dominating estimates at larger thresholds.
To control for changes in document composition, we evaluate all thresholds, models, and quantization formats on the same subset of documents eligible at the largest threshold. 
Each chunk is passed individually to the model and the generated outputs are concatenated in their original order to reconstruct the document translation. We compute chrF++ for each reconstructed document separately and report the unweighted mean across all eligible documents. The main reason for relying on this statistical measure is that neural metrics, such as xCOMET \citep{guerreiro2024xcomet}, have a limited context window of 512 tokens for the combined source, reference, and translation text.
Further details on the chunking and evaluation, including decoding parameters and prompt templates can be found in \cref{appendix:details_chunklen_eval}.

\subsection{Inference Efficiency}
The inference efficiency benchmarks are split into two parts. We perform offline benchmarking in the first stage to evaluate the speed-ups from using hardware-accelerated GEMM kernels and reduced memory traffic in isolation. In a second stage, we study optimal chunking strategies of documents in a closed-loop online inference setting for different number of concurrent users on medium-sized documents.
In general, chunking documents into shorter parts allows parallel processing, but increases the total number of requests, which vLLM needs to schedule.
By choosing the optimal chunk length, the end-to-end latency of the full document and the throughput of documents can be improved. 
However, sentences might lose context if split into too short chunks, leading to degraded translations. 
We evaluate the trade-offs of how the chunk size affects translation quality across different languages in the final subsection.
We benchmark the models using randomly generated token IDs for both inference evaluations as input prompts and disable prefix caching to ensure comparability across model sizes and different quantization formats. 
We report the results in terms of generated tokens.

\subsubsection{Offline Benchmarking}
We first benchmark the inference efficiency of the quantization formats via offline benchmarking.
While vLLM can handle requests of different lengths efficiently via continuous batching, we measure the inference statistics for fixed input and output sequence lengths and enforce the number of output tokens to the number of input tokens by ignoring the $\text{eos}_\text{token}$ for simplicity.
This holds approximately true for MT tasks, but also highly depends on the language pair and the tokenizer used \citep{martins2025eurollm}. 
We measure the wall time and output token throughput for a range of different sequence lengths and batch sizes up to 1024.

The roofline model is a useful tool to upper bound the benefits of different quantization formats for inference efficiency by putting the compute capacity and memory-bandwidth into relation.
A full derivation of the roofline model and upper bound estimates on speed-ups can be found in \cref{appendix:roofline_model}.
During low-batch decoding, inference is generally dominated by streaming model weights from memory, such that reducing weight precision can provide substantial benefits. As the batch size increases, weights are reused across more tokens and arithmetic intensity increases, reducing the benefit of weight-only quantization as projection and MLP operations become compute-bound. In this regime, activation quantization becomes important because it enables hardware-accelerated low-precision computation. Long-context decoding introduces an additional limitation: attention and KV-cache traffic grow with context length but are unaffected by weight quantization, causing the attainable end-to-end speed-up to decrease. These predictions represent idealized upper bounds and do not account for dynamic quantization overhead, kernel utilization, or scheduling and KV-cache capacity effects during online serving.

\subsubsection{Closed-Loop Online Document-Level Benchmark}

To benchmark the inference efficiency of quantized models more realistically, we perform a controlled closed-loop synthetic online benchmark, which consists of an additional orchestration layer to incorporate chunking strategies of full documents and to measure E2E latencies on the document level. 
We simulate document lengths corresponding to 1200 tokens under a growing number of concurrent users of up to 500 users and different chunk sizes ranging from 50 to 400 tokens. 

\section{Results}\label{sec:experiments}

\begin{table*}[ht!]
\caption{DocHPLT scores of EuroLLM-1.7B and EuroLLM-9B across different quantization formats in terms of chrF++. Target thresh refers to the target prompt-length threshold. Mean scores of chrF++ computed across subset of $N=27$ documents of DocHPLT. Brackets show bootstrapped 0.95 confidence intervals. For more details, please refer to \cref{subsubsec:chunking_vs_translation}.}
\centering
\begin{tabular}{lr | p{3cm}p{4cm}p{4cm}}
\toprule
Model  & Format        && Evaluation (chrF++ $\uparrow$) &\\

\midrule
 && Indep-segment translation \newline (UK $\to$ EN)   & Document-level eval \newline Target thresh=100 tok  & Document-level eval \newline Target thresh=1500 tok    \\
\midrule
EuroLLM-1.7B &  BF16 & 65.7, CI[58.2, 70.6]       & 67.6, CI[59.2, 72.6]                    & 35.6, CI[28.7, 41.1] \\
& W8A8 & 58.6, CI[51.9, 64.3] & 65.3, CI [57.2, 70.3] & 30.7, CI [24.5, 36.0]\\
& W4A16 & 63.6, CI[56.5, 69.0] & 66.2 , CI[58.7, 70.6] & 6.3, CI[4.9, 8.5] \\

\midrule
 &&  & Target thresh=100 tok  & Target thresh=4000 tok  \\

\midrule

EuroLLM-9B & BF16 & 65.2, CI[60.3, 68.8] & 69.6, CI[65.1, 72.4] & 65.23, CI[60.0, 70.12] \\
& W8A8 & 63.4, CI[58.4, 67.1]  & 68.3, CI[63.4, 71.2] & 60.22, CI[52.2, 66.2]\\
& W4A16 & 61.3, CI[56.2, 65.4] & 64.8, CI[60.5, 68.0] & 45.6, CI[ 36.8, 54.4]\\ 
\midrule

\bottomrule
\end{tabular}
\label{tab:quantization_chunklen_results}
\end{table*}

\subsection{Offline Benchmarking}

\begin{figure*}[ht!]
    \centering
    \includegraphics[width=0.94\linewidth]{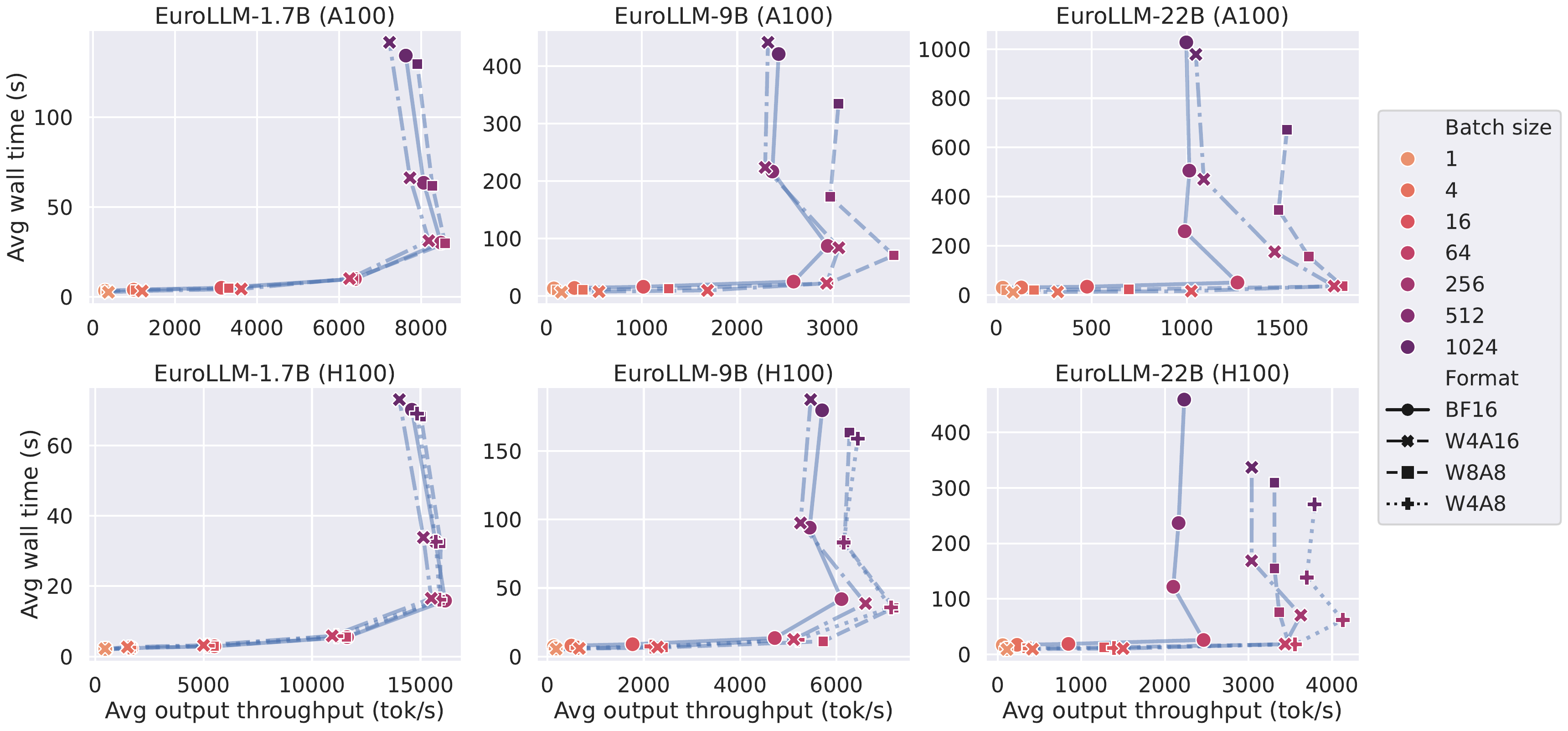}
    \caption{Offline benchmark measuring average output throughput in tok/s against average wall time for a sequence length of 1000 tokens across three sizes regimes on a A100 ({top row}) and a H100 ({bottom row}). Further results on other sequence lengths can be found in {\cref{appendix:results_offline_benchmarking}}.}
    \label{fig:offline_benchmark_pareto_curve_overview_1000}
\end{figure*}

The results of the offline benchmarks can be found in {\cref{fig:offline_benchmark_pareto_curve_overview_1000}} with further results in {\cref{appendix:results_offline_benchmarking}}. 
Across the three size regimes we observe that quantization formats have marginal benefits for small model sizes < 2B parameters.
For medium-sized models of around 9B, W4A8 and W8A8 show efficiency gains across all batch sizes, while W4A16 becomes as fast as BF16 at large batch sizes as the model becomes compute-bound. 
Finally for large MT models of 22B, we observe a difference between the H100 and A100 due to the higher compute-to-memory-bandwidth ratio on the H100. While H100 has a ridge point at $ \approx I_{\text{ridge}^{\text{H100}}} = 295 \, \text{FLOP/byte}$, A100 has a ridge point at $ \approx I_{\text{ridge}^{\text{A100}}} = 153\, \text{FLOP/byte}$ (see \cref{appendix:roofline_model} for more details). Thus, W4A16 remains competitive on the H100, while it saturates back to the BF16 baseline on the A100 at the highest batch sizes considered. We observe that for larger batch sizes, throughput actually can reduce, as is the case for the 22B model on A100. This observation is not predicted by a simple roofline model.

\begin{figure*}[ht!]
    \centering
    \includegraphics[width=0.94\linewidth]{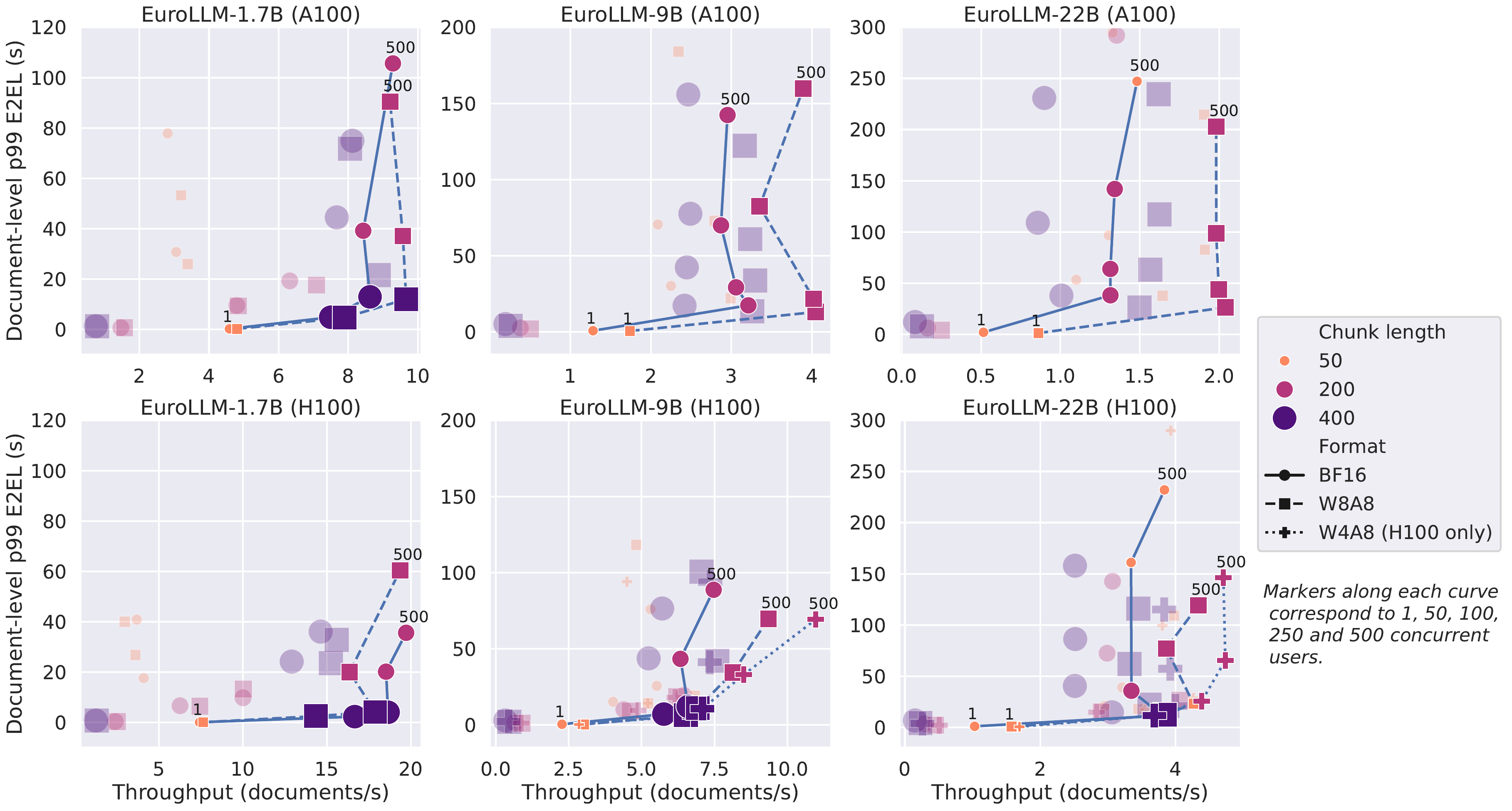}
    \caption{Closed-loop online benchmark for user concurrency from 1 to 500 users measuring throughput in documents/s against p99 full-Document latency for different chunking strategies on a document length of 1200 tokens across three model sizes on a A100 ({top row}) and a H100 ({bottom row}). Note that we did not benchmark W4A8 on the H100 for the 1.7B model size.}
    \label{fig:online_benchmark_pareto_curve_overview_EuroLLM2}
\end{figure*}

\begin{figure*}[ht!]
    \centering
    \includegraphics[width=0.7\linewidth]{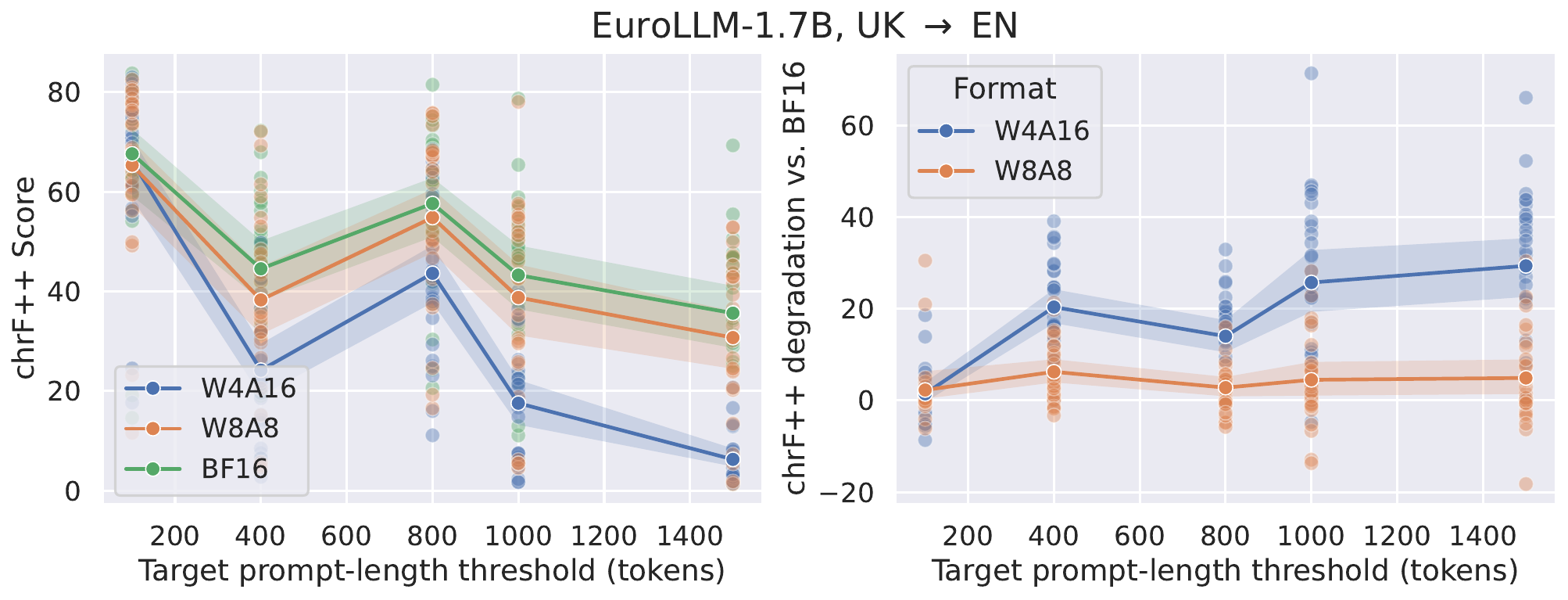}
    \includegraphics[width=0.7\linewidth]{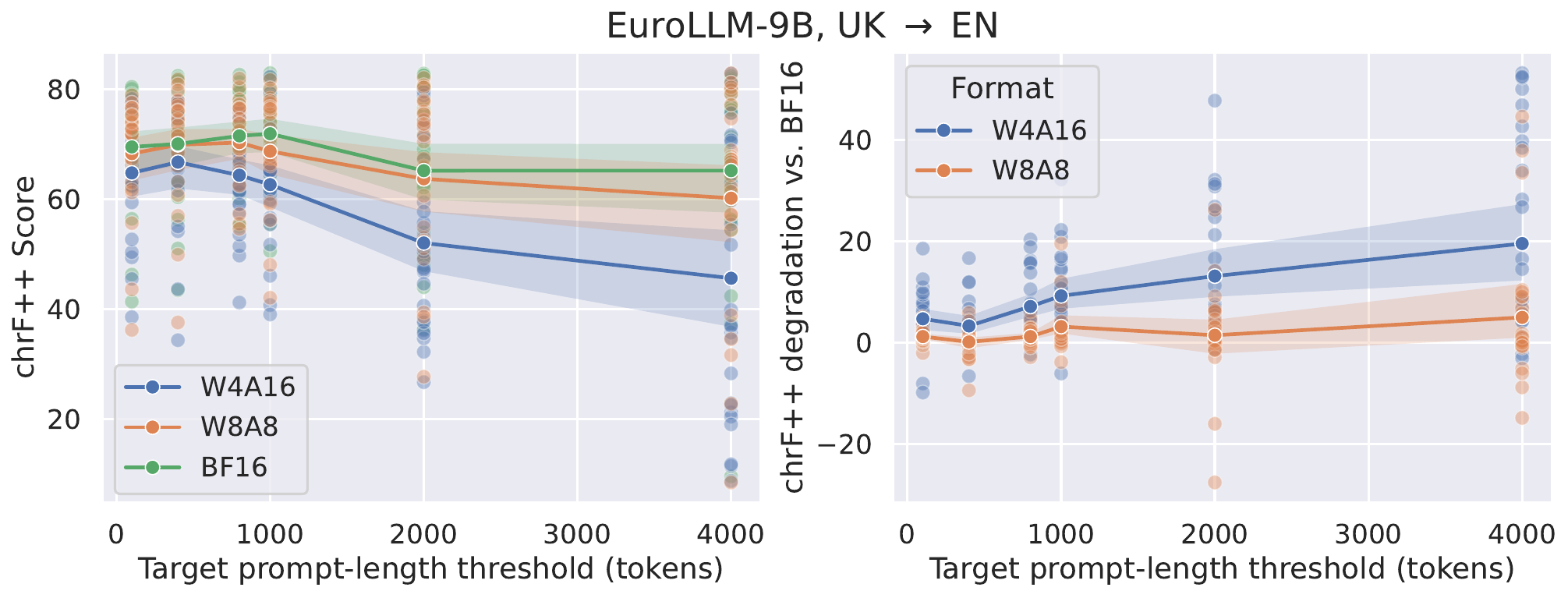}
    \caption{Chunk length vs. chrF++ Score for EuroLLM-1.7B (top row) and EuroLLM-9B (bottom row) across different quantization formats averaged across a subset of 27 documents from DocHPLT. Subplots on the right corresponds to paired per-document BF16 - quantized differences, where positive values mean degradation relative to BF16. Shaded area corresponds to bootstrapped 95\% CI of document-level means. Note the different $y$ limits for better visibility. Every point corresponds to individual document scores. Additional results on other language pairs be found in {\cref{appendix:results_chunklen_chrF}}.}
    \label{fig:chunklen_overview2}
\end{figure*}

\subsection{Closed-Loop Online Document-Level Benchmark}

Based on the offline benchmark and quantization results, we limit the comparison to BF16, W8A8, and W4A8.
The results can be found in \cref{fig:online_benchmark_pareto_curve_overview_EuroLLM2}. 
We make the following observations based on the results: Quantization is not always beneficial. 
In particular for the 1.7B model deployed on a H100, BF16 achieves higher throughput at lower latencies than W8A8 quantization. The reason for this is that the W4A8 and W8A8 schemes add overhead by applying dynamic activation quantization.
In addition, the online benchmark further shows that document chunking can provide additional efficiency gains. The Pareto-optimal chunking strategy depends on the specific deployment setting and workload. 
At low concurrency, it is always preferable to use the smallest acceptable chunk length to allow for parallel processing, while for higher concurrency the Pareto-optimal chunk length varies mostly between 200 and 400 tokens. 
At the highest considered concurrency on the 22B model under BF16 precision, a trade-off between higher throughput and lower latency is also observable. 
While shorter chunk length allow for a better KV cache management and lead to a higher throughput, the tail latency also increases considerably. 
The quantized variants under the same workload do not show this behavior. We speculate that this is because they have more memory headroom by saving 20-30 GB of model memory.
Apart from the 1.7B model deployed on a H100 GPU all other configurations provide efficiency gains by combining W4A8 or W8A8 quantization with a chunking strategy.  
While chunking of documents improves the inference efficiency, it might affect the translation quality by losing relevant context for correct translations.
We discuss the effect of chunking strategies on translation quality in the next subsection. 

\subsection{Effect of Chunking Strategy on Translation Quality}\label{subsubsec:chunking_vs_translation}

Unlike the synthetic random-token workloads used in the previous section to cleanly isolate scheduling performance, evaluating translation quality requires real-world data.
In this section we evaluate MT quality on document-level translations. Our results reveal that standard segment-level evaluation can underestimate the interaction between quantization and long-context document translation for certain quantization formats, translation directions and models.

The results for the UK $\to$ EN language pair can be found in \cref{fig:chunklen_overview2} with further results on EN $\to$ UK in \cref{appendix:results_chunklen_chrF}. 
A corresponding WMT24++ analysis, which can be found in \cref{appendix:results_chunklen_chrF} shows a weaker and less consistent effect up to a token threshold of 1000.
The figures show that even though EuroLLM-1.7B has a context length of 4096, it translates longer inputs considerably worse, even at full BF16 precision. 
This observation is considerably weaker for the BF16 checkpoint of EuroLLM-9B, which was explicitly trained for long contexts up to 32K.  
In addition, our evaluation shows that standard segment-level MT benchmarks, such as WMT \citep{deutsch2025wmt24++} or Flores+ \citep{flores101-22} can potentially underestimate the degradation of quantized MT models under long context for certain translation directions and quantization formats. 
This is highlighted in \cref{tab:quantization_chunklen_results} as well as the right side of \cref{fig:chunklen_overview2}, in which the W4A16 quantized checkpoints of EuroLLM-1.7B and EuroLLM-9B suggest a lower degradation on segment-level translations than what is observed if longer chunks are being translated directly. 
We note however that this effect is considerably weaker for W8A8 up to the considered sequence length as well as for the reverse translation direction EN $\to$ UK, as can be seen in {\cref{fig:chunklen_overview_EN_UK}}.

\section{Discussion}\label{sec:discussion}

 The closed-loop online benchmarks have shown that document chunking can improve model serving efficiency. 
Across model sizes, mainly medium-sized and larger MT models benefit from quantization, while small models < 2B only show marginal benefits or even slight slow down compared to the BF16 baseline due to the overhead introduced by dynamic activation quantization.
The efficiency gains across the full range of workloads arise from a combination of higher compute capacity by making use of INT8 tensor cores and reducing the model size, where the latter is beneficial both at low workloads when inference is memory bandwidth-bound as well as during capacity-bound serving under high user concurrency where memory savings benefit KV cache management and scheduling efficiency during online inference. \\
Furthermore, our experiments on how chunk length affects translation quality have shown that standard segment-level MT benchmarks can potentially underestimate the degradation for longer context translation for certain translation directions and quantization formats, which needs to be carefully evaluated case-by-case.
While in some cases the quantized checkpoint follows the BF16 baseline closely, it can lead to a measurable degradation in MT quality in other cases, such as the W4A16 quantized checkpoints of EuroLLM-1.7B and EuroLLM-9B evaluated on UK $\to$ EN.
We hypothesize that the different context length ranging from 4K to 32K of the models considered as well as the data that the models have been trained on play a major role.   
For instance, EuroLLM-1.7B is instruction-tuned only on segment-level translation pairs \citep{martins2025eurollm}.
In addition, the sentence-level calibration data of OPUS \citep{tiedemann2016opus} could have biased the quantized model checkpoints toward shorter translation thresholds.
Although identifying the mechanistic reason underlying the sensitivity induced by quantization is still open and out of scope for this work, we hypothesize that a short Quantization-aware training (QAT) or distillation run with long context translation samples can help to close this gap. 
If the Pareto-optimal chunk length in terms of efficiency is longer than for translation quality, one could resolve this discrepancy by simulating shorter chunks using appropriate block-diagonal attention masks (though this adds additional overhead), but in the other case it results in a trade-off between achieving higher inference efficiency and better translation quality. 
We note however that we do not construct a joint quality–efficiency Pareto frontier in this work, since translation quality is evaluated only for a subset of the models and quantization formats considered in the inference benchmarks.

We summarize key takeaways of our work below:
\begin{enumerate}
    \item \textbf{Quantization to W8A8 or W4A8 can improve the latency-throughput tradeoff, depending on hardware and model size.}
    Among the configurations supported in our vLLM setup, for models of 9B parameters and above, W8A8 on the A100 and W4A8 on the H100 can provide improved throughput at similar latency. At higher batch sizes, activation quantization is particularly important on the A100, whose lower compute-to-memory-bandwidth ratio causes inference to become compute-bound sooner.

    \item \textbf{Quantization and document chunking should be optimized jointly for inference efficiency.}
    Across most evaluated deployment settings, combining W8A8 or W4A8 with chunk lengths of approximately 200--400 tokens improves the latency-throughput Pareto frontier for medium-length documents.

    \item \textbf{Sentence-level MT benchmarks can underestimate long-context degradation.}
    Evaluations based on isolated segments can underestimate how translation quality changes with longer context and can therefore conceal degradation that occurs during document-level inference, in particular when deploying quantized models. This strongly depends on the considered quantization format and translation direction.

\end{enumerate}

\section{Limitations and Future Work}\label{sec:limitations}

In this work, we only considered GPTQ \citep{frantar2022gptq} combined with \citep{xiao2023smoothquant} as a quantization method and it is possible that more sophisticated PTQ methods or QAT do not show degradation for longer contexts observed on the quantized EuroLLM models. We also did not control for the calibration data used for PTQ, and it remains to be verified whether longer context calibration data can reduce the degradation.  
We note that we apply dynamic activation quantization, which added an overhead in the inference benchmark, which could potentially be removed. While static activation quantization can help, it risks to potentially degrade the translation quality. 
Further, we point out that W4A8 was not tested on A100 due to missing vLLM support. It thus remains an open question, whether W4A8 will be preferable over W8A8 once corresponding software-support exists. 
Furthermore, we only evaluated how translation quality changes with chunk lengths on two translation directions, so they might be different for low-resource languages. \\
Our quality evaluation was also performed only across a subset of 27 UK $\to$ EN, respectively 13 EN $\to$ UK documents of DocHPLT, which needs to be extended to more documents in future work.
While we ran quality experiments up to a threshold of 4K tokens, it is still not close to the 32K context length of EuroLLM-9B, so we were not able to characterize the behavior of EuroLLM-9B close to its context limit.
Finally, in this work we focused only on weight and activation quantization. We leave the consideration of KV cache quantization for inference and quality retention for future work. 

\section{Conclusion}\label{sec:conclusion}

In this work, we have studied quantization trade-offs for efficient inference deployment for machine translation on the EuroLLM model family on a A100 and a H100 GPU. 
By evaluating the inference efficiency across three models, spanning 1.7B to 22B under closed-loop synthetic online benchmarks on vLLM, we find that for model sizes of 9B and above W8A8 generally provides the strongest Pareto performance in the evaluated configurations on the A100, while W4A8 provides the strongest Pareto performance on the H100. 
For small models < 2B W8A8 leads to marginal efficiency gains or even slight slowdown.
In our synthetic document-serving benchmark we show that chunking medium-sized documents into chunks of 200-400 tokens is Pareto-optimal for inference efficiency. Finally, we observe that translation quality is better for shorter thresholds, while the robustness for longer sequences differs for model sizes, quantization format, and translation direction.
\newpage

\bibliography{references}

@article{flores101-22,
    title = {The {F}lores-101 Evaluation Benchmark for Low-Resource and Multilingual Machine Translation},
    author = {Goyal, Naman and Gao, Cynthia and Chaudhary, Vishrav and Chen, Peng-Jen and Wenzek, Guillaume and Ju, Da and Krishnan, Sanjana and Ranzato, Marc’Aurelio and Guzmán, Francisco and Fan, Angela},
    journal = {Transactions of the Association for Computational Linguistics},
    volume = {10},
    year = {2022},
}

@article{abecassis2025pretraining,
  title={Pretraining large language models with nvfp4},
  author={Abecassis, Felix and Agrusa, Anjulie and Ahn, Dong and Alben, Jonah and Alborghetti, Stefania and Andersch, Michael and Arayandi, Sivakumar and Bjorlin, Alexis and Blakeman, Aaron and Briones, Evan and others},
  journal={arXiv preprint arXiv:2509.25149},
  year={2025}
}

@article{achiam2023gpt,
  title={Gpt-4 technical report},
  author={Achiam, Josh and Adler, Steven and Agarwal, Sandhini and Ahmad, Lama and Akkaya, Ilge and Aleman, Florencia Leoni and Almeida, Diogo and Altenschmidt, Janko and Altman, Sam and Anadkat, Shyamal and others},
  journal={arXiv preprint arXiv:2303.08774},
  year={2023}
}

@inproceedings{agrawal2024taming,
  title={Taming $\{$Throughput-Latency$\}$ tradeoff in $\{$LLM$\}$ inference with $\{$Sarathi-Serve$\}$},
  author={Agrawal, Amey and Kedia, Nitin and Panwar, Ashish and Mohan, Jayashree and Kwatra, Nipun and Gulavani, Bhargav and Tumanov, Alexey and Ramjee, Ramachandran},
  booktitle={18th USENIX symposium on operating systems design and implementation (OSDI 24)},
  pages={117--134},
  year={2024}
}

@article{cheng2025seed,
  title={Seed-x: Building strong multilingual translation llm with 7b parameters},
  author={Cheng, Shanbo and Bao, Yu and Cao, Qian and Huang, Luyang and Kang, Liyan and Liu, Zhicheng and Lu, Yu and Zhu, Wenhao and Chen, Jingwen and Huang, Zhichao and others},
  journal={arXiv preprint arXiv:2507.13618},
  year={2025}
}

@inproceedings{coignion2024performance,
  title={A performance study of llm-generated code on leetcode},
  author={Coignion, Tristan and Quinton, Cl{\'e}ment and Rouvoy, Romain},
  booktitle={Proceedings of the 28th international conference on evaluation and assessment in software engineering},
  pages={79--89},
  year={2024}
}

@inproceedings{deutsch2025wmt24++,
  title={WMT24++: Expanding the language coverage of WMT24 to 55 languages \& dialects},
  author={Deutsch, Daniel and Briakou, Eleftheria and Caswell, Isaac Rayburn and Finkelstein, Mara and Galor, Rebecca and Juraska, Juraj and Kovacs, Geza and Lui, Alison and Rei, Ricardo and Riesa, Jason and others},
  booktitle={Findings of the Association for Computational Linguistics: ACL 2025},
  pages={12257--12284},
  year={2025}
}

@article{frantar2022gptq,
  title={Gptq: Accurate post-training quantization for generative pre-trained transformers},
  author={Frantar, Elias and Ashkboos, Saleh and Hoefler, Torsten and Alistarh, Dan},
  journal={arXiv preprint arXiv:2210.17323},
  year={2022}
}

@article{guerreiro2024xcomet,
  title={xcomet: Transparent machine translation evaluation through fine-grained error detection},
  author={Guerreiro, Nuno M and Rei, Ricardo and Stigt, Daan van and Coheur, Luisa and Colombo, Pierre and Martins, Andr{\'e} FT},
  journal={Transactions of the Association for Computational Linguistics},
  volume={12},
  pages={979--995},
  year={2024},
  publisher={MIT Press 255 Main Street, 9th Floor, Cambridge, Massachusetts 02142, USA~…}
}

@article{hooper2024kvquant,
  title={Kvquant: Towards 10 million context length llm inference with kv cache quantization},
  author={Hooper, Coleman and Kim, Sehoon and Mohammadzadeh, Hiva and Mahoney, Michael W and Shao, Yakun S and Keutzer, Kurt and Gholami, Amir},
  journal={Advances in Neural Information Processing Systems},
  volume={37},
  pages={1270--1303},
  year={2024}
}

@inproceedings{jarmusch2026microbenchmarking,
  title={Microbenchmarking nvidia’s blackwell architecture: An in-depth architectural analysis},
  author={Jarmusch, Aaron and Chandrasekaran, Sunita},
  booktitle={2026 IEEE International Parallel and Distributed Processing Symposium (IPDPS)},
  pages={1026--1036},
  year={2026},
  organization={IEEE}
}

@inproceedings{karpinska-iyyer-2023-large,
    title = "Large Language Models Effectively Leverage Document-level Context for Literary Translation, but Critical Errors Persist",
    author = "Karpinska, Marzena  and
      Iyyer, Mohit",
    editor = "Koehn, Philipp  and
      Haddow, Barry  and
      Kocmi, Tom  and
      Monz, Christof",
    booktitle = "Proceedings of the Eighth Conference on Machine Translation",
    month = dec,
    year = "2023",
    address = "Singapore",
    publisher = "Association for Computational Linguistics",
    url = "https://aclanthology.org/2023.wmt-1.41/",
    doi = "10.18653/v1/2023.wmt-1.41",
    pages = "419--451",
}

@inproceedings{kwon2023efficient,
  title={Efficient memory management for large language model serving with pagedattention},
  author={Kwon, Woosuk and Li, Zhuohan and Zhuang, Siyuan and Sheng, Ying and Zheng, Lianmin and Yu, Cody Hao and Gonzalez, Joseph and Zhang, Hao and Stoica, Ion},
  booktitle={Proceedings of the 29th symposium on operating systems principles},
  pages={611--626},
  year={2023}
}

@inproceedings{laubli-etal-2018-machine,
    title = "Has Machine Translation Achieved Human Parity? A Case for Document-level Evaluation",
    author = {L{\"a}ubli, Samuel  and
      Sennrich, Rico  and
      Volk, Martin},
    editor = "Riloff, Ellen  and
      Chiang, David  and
      Hockenmaier, Julia  and
      Tsujii, Jun{'}ichi",
    booktitle = "Proceedings of the 2018 Conference on Empirical Methods in Natural Language Processing",
    month = oct # "-" # nov,
    year = "2018",
    address = "Brussels, Belgium",
    publisher = "Association for Computational Linguistics",
    url = "https://aclanthology.org/D18-1512/",
    doi = "10.18653/v1/D18-1512",
    pages = "4791--4796",
}

@article{lin2025awq,
  title={Awq: Activation-aware weight quantization for on-device llm compression and acceleration},
  author={Lin, Ji and Tang, Jiaming and Tang, Haotian and Yang, Shang and Xiao, Guangxuan and Han, Song},
  journal={GetMobile: Mobile Computing and Communications},
  volume={28},
  number={4},
  pages={12--17},
  year={2025},
  publisher={ACM New York, NY, USA}
}

@article{li2024evaluating,
  title={Evaluating quantized large language models},
  author={Li, Shiyao and Ning, Xuefei and Wang, Luning and Liu, Tengxuan and Shi, Xiangsheng and Yan, Shengen and Dai, Guohao and Yang, Huazhong and Wang, Yu},
  journal={arXiv preprint arXiv:2402.18158},
  year={2024}
}

@article{liu2025generalist,
  title={A generalist medical language model for disease diagnosis assistance},
  author={Liu, Xiaohong and Liu, Hao and Yang, Guoxing and Jiang, Zeyu and Cui, Shuguang and Zhang, Zhaoze and Wang, Huan and Tao, Liyuan and Sun, Yongchang and Song, Zhu and others},
  journal={Nature medicine},
  volume={31},
  number={3},
  pages={932--942},
  year={2025},
  publisher={Nature Publishing Group US New York}
}

@inproceedings{lopes-etal-2020-document,
    title = "Document-level Neural {MT}: A Systematic Comparison",
    author = "Lopes, Ant{\'o}nio  and
      Farajian, M. Amin  and
      Bawden, Rachel  and
      Zhang, Michael  and
      Martins, Andr{\'e} F. T.",
    editor = "Martins, Andr{\'e}  and
      Moniz, Helena  and
      Fumega, Sara  and
      Martins, Bruno  and
      Batista, Fernando  and
      Coheur, Luisa  and
      Parra, Carla  and
      Trancoso, Isabel  and
      Turchi, Marco  and
      Bisazza, Arianna  and
      Moorkens, Joss  and
      Guerberof, Ana  and
      Nurminen, Mary  and
      Marg, Lena  and
      Forcada, Mikel L.",
    booktitle = "Proceedings of the 22nd Annual Conference of the European Association for Machine Translation",
    month = nov,
    year = "2020",
    address = "Lisboa, Portugal",
    publisher = "European Association for Machine Translation",
    url = "https://aclanthology.org/2020.eamt-1.24/",
    pages = "225--234",
  
}

@article{ma2024era,
  title={The era of 1-bit llms: All large language models are in 1.58 bits},
  author={Ma, Shuming and Wang, Hongyu and Ma, Lingxiao and Wang, Lei and Wang, Wenhui and Huang, Shaohan and Dong, Li and Wang, Ruiping and Xue, Jilong and Wei, Furu},
  journal={arXiv preprint arXiv:2402.17764},
  year={2024}
}

@inproceedings{marchisio2024does,
  title={How does quantization affect multilingual LLMs?},
  author={Marchisio, Kelly and Dash, Saurabh and Chen, Hongyu and Aumiller, Dennis and {\"U}st{\"u}n, Ahmet and Hooker, Sara and Ruder, Sebastian},
  booktitle={Findings of the Association for Computational Linguistics: EMNLP 2024},
  pages={15928--15947},
  year={2024}
}

@article{marie2025uneven,
  title={The Uneven Impact of Post-Training Quantization in Machine Translation},
  author={Marie, Benjamin and Fujita, Atsushi},
  journal={arXiv preprint arXiv:2508.20893},
  year={2025}
}

@article{martins2025eurollm,
  title={Eurollm: Multilingual language models for europe},
  author={Martins, Pedro Henrique and Fernandes, Patrick and Alves, Jo{\~a}o and Guerreiro, Nuno M and Rei, Ricardo and Alves, Duarte M and Pombal, Jos{\'e} and Farajian, Amin and Faysse, Manuel and Klimaszewski, Mateusz and others},
  journal={Procedia Computer Science},
  volume={255},
  pages={53--62},
  year={2025},
  publisher={Elsevier}
}

@article{maruf2021survey,
  title={A survey on document-level neural machine translation: Methods and evaluation},
  author={Maruf, Sameen and Saleh, Fahimeh and Haffari, Gholamreza},
  journal={ACM Computing Surveys (CSUR)},
  volume={54},
  number={2},
  pages={1--36},
  year={2021},
  publisher={ACM New York, NY, USA}
}

@article{maskey20261,
  title={1-Bit Wonder: Improving QAT Performance in the Low-Bit Regime through K-Means Quantization},
  author={Maskey, Sohir and Eichenberg, Constantin and Messner, Johannes and Orr, Douglas},
  journal={arXiv preprint arXiv:2602.15563},
  year={2026}
}

@inproceedings{nagel2019data,
  title={Data-free quantization through weight equalization and bias correction},
  author={Nagel, Markus and Baalen, Mart van and Blankevoort, Tijmen and Welling, Max},
  booktitle={Proceedings of the IEEE/CVF international conference on computer vision},
  pages={1325--1334},
  year={2019}
}

@misc{nvidia2023tensorrtllm,
  author       = {{NVIDIA Corporation}},
  title        = {{TensorRT-LLM: An open-source library for optimizing LLM inference}},
  year         = {2023},
  howpublished = {\url{https://github.com/NVIDIA/TensorRT-LLM}},
  note         = {Accessed: 2026-07-25}
}

@article{ouyang2022training,
  title={Training language models to follow instructions with human feedback},
  author={Ouyang, Long and Wu, Jeff and Jiang, Xu and Almeida, Diogo and Wainwright, Carroll L and Mishkin, Pamela and Zhang, Chong and Agarwal, Sandhini and Slama, Katarina and Ray, Alex and others},
  journal={arXiv preprint arXiv:2203.02155},
  year={2022}
}

@inproceedings{o2025dochplt,
  title={DocHPLT: A massively multilingual document-level translation dataset},
  author={O’Brien, Dayy{\'a}n and Malik, Bhavitvya and de Gibert, Ona and Chen, Pinzhen and Haddow, Barry and Tiedemann, J{\"o}rg},
  booktitle={Proceedings of the Tenth Conference on Machine Translation},
  pages={286--300},
  year={2025}
}

@inproceedings{popovic2017chrf++,
  title={chrF++: words helping character n-grams},
  author={Popovi{\'c}, Maja},
  booktitle={Proceedings of the second conference on machine translation},
  pages={612--618},
  year={2017}
}

@inproceedings{qu2025semantic,
  title={Is semantic chunking worth the computational cost?},
  author={Qu, Renyi and Tu, Ruixuan and Bao, Forrest},
  booktitle={Findings of the Association for Computational Linguistics: NAACL 2025},
  pages={2155--2177},
  year={2025}
}

@article{rouhani2023microscaling,
  title={Microscaling data formats for deep learning},
  author={Rouhani, Bita Darvish and Zhao, Ritchie and More, Ankit and Hall, Mathew and Khodamoradi, Alireza and Deng, Summer and Choudhary, Dhruv and Cornea, Marius and Dellinger, Eric and Denolf, Kristof and others},
  journal={arXiv preprint arXiv:2310.10537},
  year={2023}
}

@article{su2024roformer,
  title={Roformer: Enhanced transformer with rotary position embedding},
  author={Su, Jianlin and Ahmed, Murtadha and Lu, Yu and Pan, Shengfeng and Bo, Wen and Liu, Yunfeng},
  journal={Neurocomputing},
  volume={568},
  pages={127063},
  year={2024},
  publisher={Elsevier}
}

@inproceedings{tiedemann2016opus,
  title={OPUS--parallel corpora for everyone},
  author={Tiedemann, J{\"o}rg},
  booktitle={Proceedings of the 19th Annual Conference of the European Association for Machine Translation: Projects/Products},
  year={2016}
}

@inproceedings{wang-etal-2023-document-level,
    title = "Document-Level Machine Translation with Large Language Models",
    author = "Wang, Longyue  and
      Lyu, Chenyang  and
      Ji, Tianbo  and
      Zhang, Zhirui  and
      Yu, Dian  and
      Shi, Shuming  and
      Tu, Zhaopeng",
    editor = "Bouamor, Houda  and
      Pino, Juan  and
      Bali, Kalika",
    booktitle = "Proceedings of the 2023 Conference on Empirical Methods in Natural Language Processing",
    month = dec,
    year = "2023",
    address = "Singapore",
    publisher = "Association for Computational Linguistics",
    url = "https://aclanthology.org/2023.emnlp-main.1036/",
    doi = "10.18653/v1/2023.emnlp-main.1036",
    pages = "16646--16661",
}

@article{wang2023bitnet,
  title={Bitnet: Scaling 1-bit transformers for large language models},
  author={Wang, Hongyu and Ma, Shuming and Dong, Li and Huang, Shaohan and Wang, Huaijie and Ma, Lingxiao and Yang, Fan and Wang, Ruiping and Wu, Yi and Wei, Furu},
  journal={arXiv preprint arXiv:2310.11453},
  year={2023}
}

@article{wu2025bitnet,
  title={BitNet Distillation},
  author={Wu, Xun and Huang, Shaohan and Wang, Wenhui and Song, Ting and Dong, Li and Xia, Yan and Wei, Furu},
  journal={arXiv preprint arXiv:2510.13998},
  year={2025}
}

@inproceedings{xiao2023smoothquant,
  title={Smoothquant: Accurate and efficient post-training quantization for large language models},
  author={Xiao, Guangxuan and Lin, Ji and Seznec, Mickael and Wu, Hao and Demouth, Julien and Han, Song},
  booktitle={International conference on machine learning},
  pages={38087--38099},
  year={2023},
  organization={PMLR}
}

@article{xin2026quantization,
  title={Quantization-Aware Distillation for NVFP4 Inference Accuracy Recovery},
  author={Xin, Meng and Priyadarshi, Sweta and Xin, Jingyu and Kartal, Bilal and Vavre, Aditya and Thekkumpate, Asma Kuriparambil and Chen, Zijia and Mahabaleshwarkar, Ameya Sunil and Shahaf, Ido and Bercovich, Akhiad and others},
  journal={arXiv preprint arXiv:2601.20088},
  year={2026}
}

@inproceedings{yu2022orca,
  title={Orca: A distributed serving system for $\{$Transformer-Based$\}$ generative models},
  author={Yu, Gyeong-In and Jeong, Joo Seong and Kim, Geon-Woo and Kim, Soojeong and Chun, Byung-Gon},
  booktitle={16th USENIX symposium on operating systems design and implementation (OSDI 22)},
  pages={521--538},
  year={2022}
}

@inproceedings{zhang2024integer,
  title={Integer or floating point? new outlooks for low-bit quantization on large language models},
  author={Zhang, Yijia and Zhao, Lingran and Cao, Shijie and Zhang, Sicheng and Wang, Wenqiang and Cao, Ting and Yang, Fan and Yang, Mao and Zhang, Shanghang and Xu, Ningyi},
  booktitle={2024 IEEE International Conference on Multimedia and Expo (ICME)},
  pages={1--6},
  year={2024},
  organization={IEEE}
}

@article{zhang2024qqq,
  title={Qqq: Quality quattuor-bit quantization for large language models},
  author={Zhang, Ying and Zhang, Peng and Huang, Mincong and Xiang, Jingyang and Wang, Yujie and Wang, Chao and Zhang, Yineng and Yu, Lei and Liu, Chuan and Lin, Wei},
  journal={arXiv preprint arXiv:2406.09904},
  year={2024}
}

@article{zheng2026hy,
  title={Hy-MT2: A Family of Fast, Efficient and Powerful Multilingual Translation Models in the Wild},
  author={Zheng, Mao and Li, Zheng and Chen, Tao and Lv, Bo and Sun, Mingrui and Song, Mingyang and Song, Jinlong and Huang, Hong and Wu, Decheng and Wang, Hai and others},
  journal={arXiv preprint arXiv:2605.22064},
  year={2026}
}

\appendix

\newpage

\section{Appendix}

\subsection{A concrete GEMM-only roofline model}\label{appendix:roofline_model}

We first consider an idealized model that includes only GEMM FLOPs and the
movement of GEMM operands between HBM and the GPU. For
\[
    Y = XW,
    \qquad
    X \in \mathbb{R}^{M\times K},
    \quad
    W \in \mathbb{R}^{K\times N},
\]
the number of FLOPs and transferred bytes are approximated by
\[
\begin{aligned}
    F_{\mathrm{gemm}} & = 2MKN,
    \\
    Q_{\mathrm{gemm}}
    & =
    b_a MK + b_w KN + b_a MN,
\end{aligned}
\]
where \(b_a\) and \(b_w\) denote the number of bytes per activation and
weight, respectively. The arithmetic intensity is therefore
\[
    I_{\mathrm{gemm}}
    =
    \frac{2MKN}
         {b_a MK+b_w KN+b_a MN}.
\]

For a square hidden projection with \(K=N=d\), this becomes
\begin{equation}
    I_{\mathrm{gemm}}(M;d)
    =
    \frac{2Md^2}
         {b_wd^2+2b_aMd}.
    \label{eq:gemm-intensity}
\end{equation}
For \(M\ll d\), weight traffic dominates and
\[
    I_{\mathrm{gemm}}(M;d)
    \approx
    \frac{2M}{b_w}.
\]
Consequently, with BF16 activations, the ideal arithmetic intensity is
approximately \(M\), \(2M\), and \(4M\) FLOP/byte for BF16, INT8, and INT4
weights, respectively.

We instantiate the analysis for the three EuroLLM models, see \Cref{tab:eurollm} for their architectural details.

\begin{table*}[!t]
\centering
\caption{Architectural details of EuroLLM models.}
\label{tab:eurollm}
\begin{tabular}{lrrrrrr}
\toprule
Model
& \(L\)
& \(d\)
& \(d_{\mathrm{ff}}\)
& \(n_q/n_{\mathrm{kv}}\)
& \(d_{\mathrm{kv}}\)
& Parameters \\
\midrule
EuroLLM-1.7B
& 24 & 2048 & 5632
& \(16/8\) & 1024 & \(1.657\)B \\
EuroLLM-9B
& 42 & 4096 & 12288
& \(32/8\) & 1024 & \(9.152\)B \\
EuroLLM-22B
& 54 & 6144 & 16384
& \(48/8\) & 1024 & \(22.637\)B \\
\bottomrule
\end{tabular}
\end{table*}

The roofline model predicts the achievable GEMM performance, measured in FLOP/s, as
\[
P_{\mathrm{gemm}}
=
\min\left(
P_{\mathrm{peak}},
\mathrm{BW}_{\mathrm{HBM}} I_{\mathrm{gemm}}
\right),
\]
where $P_{\mathrm{peak}}$ denotes the hardware's theoretical peak compute throughput in FLOP/s and $\mathrm{BW}_{\mathrm{HBM}}$ its peak HBM bandwidth in bytes/s. The product $\mathrm{BW}_{\mathrm{HBM}} I_{\mathrm{gemm}}$ is the maximum compute throughput that can be sustained by the available memory bandwidth.

Using dense BF16 Tensor Core throughput, an H100 SXM has an approximate
ridge point of
\[
    I_{\mathrm{ridge}}^{\mathrm{H100}}
    =
    \frac{989\ \mathrm{TFLOP/s}}
         {3.35\ \mathrm{TB/s}}
    \approx 295\ \mathrm{FLOP/byte},
\]
while an A100 80GB SXM has
\[
    I_{\mathrm{ridge}}^{\mathrm{A100}}
    =
    \frac{312\ \mathrm{TFLOP/s}}
         {2.039\ \mathrm{TB/s}}
    \approx 153\ \mathrm{FLOP/byte}.
\]
Here we assume that weight-only quantized kernels unpack and dequantize
weights on-chip and use the BF16 Tensor Core compute ceiling. This is an
optimistic upper bound.

With BF16 activations, the approximate number of GEMM rows (calculated via \Cref{eq:gemm-intensity}) required to reach
the compute-bound regime are shown in \Cref{tab:eurollm-gemm-rows}.
\begin{table*}[!t]
\centering
\caption{Approximate number of GEMM rows required to reach the compute-bound regime under BF16 activations, calculated via \cref{eq:gemm-intensity}.}
\label{tab:eurollm-gemm-rows}
\begin{tabular}{llrrr}
\toprule
Model & GPU & BF16 weights & INT8 weights & INT4 weights \\
\midrule
EuroLLM-1.7B & H100 & \(415\) & \(207\) & \(104\) \\
              & A100 & \(180\) & \(90\)  & \(45\)  \\
\midrule
EuroLLM-9B   & H100 & \(345\) & \(172\) & \(86\)  \\
              & A100 & \(165\) & \(83\)  & \(41\)  \\
\midrule
EuroLLM-22B  & H100 & \(327\) & \(163\) & \(82\)  \\
              & A100 & \(161\) & \(81\)  & \(40\)  \\
\bottomrule
\end{tabular}
\end{table*}

During decoding, \(M=B\), where \(B\) is the number of tokens processed
together. At batch size one, all three EuroLLM models are therefore strongly
memory-bandwidth bound. In this ideal model, INT8 and INT4 weights reduce
weight traffic by factors of two and four and provide corresponding speedups.

As the batch size increases, the same weights are reused across more tokens.
For example, the INT4 projection GEMMs become nominally compute-bound on H100
at batch sizes of approximately \(104\), \(86\), and \(82\) for EuroLLM-1.7B,
EuroLLM-9B, and EuroLLM-22B, respectively. Beyond this point, further reducing
weight traffic does not improve the idealized GEMM runtime.

During prefill, \(M=BS\), where \(S\) is the prompt sequence length.
Consequently, even at batch size one, the projection and MLP GEMMs become
compute-bound after only a few hundred prompt tokens. The GEMM-only model
therefore predicts that weight-only quantization is most useful for low-batch
decoding and much less useful for long-sequence prefill.

These theoretical considerations also generalize directly to activation quantization, provided that the hardware and kernels support computation in the chosen activation format. The effect of activation precision can be modeled by substituting the corresponding number of activation bytes into $b_a$ in \cref{eq:gemm-intensity} and using the peak compute throughput of the corresponding hardware format. For example, if INT8 computation provides twice the peak throughput of BF16 while HBM bandwidth remains unchanged, the hardware ridge point doubles. Consequently, the GEMM remains memory-bound up to approximately twice the batch size, allowing further weight quantization to provide speedups over a correspondingly larger range of batch sizes. Reducing the activation size from two bytes to one byte further decreases activation traffic, although this effect is small when weight traffic dominates.

\paragraph{Why decode can nevertheless benefit less than predicted.}

The GEMM-only model predicts no dependence of decode GEMMs on the existing
context length. The projection and MLP GEMMs have \(M=B\), irrespective of
whether the context contains \(1{,}000\) or \(30{,}000\) tokens. Thus, under
this model, decode latency would be constant with respect to context length.
The observed context-length dependence comes primarily from attention and its
KV-cache traffic.

For a model with \(L\) layers and KV dimension \(d_{\mathrm{kv}}\), reading
the BF16 key and value cache during one decoding step requires approximately
\[
    Q_{\mathrm{KV}}
    =
    2BLSd_{\mathrm{kv}}b_a,
    \label{eq:kv-bytes}
\]
where the factor two accounts for keys and values. Unlike weight traffic,
this traffic grows linearly with both batch size and context length.

The corresponding attention FLOPs are approximately
\[
    F_{\mathrm{attn}}
    \approx
    4BLSd,
\]
giving an idealized attention arithmetic intensity of
\[
    I_{\mathrm{attn}}
    \approx
    \frac{2d}
         {b_a d_{\mathrm{kv}}}
    =
    \frac{2g}{b_a},
    \qquad
    g=\frac{d}{d_{\mathrm{kv}}}.
\]

All three EuroLLM models have eight KV heads with head dimension \(128\), and
therefore \(d_{\mathrm{kv}}=1024\). Their GQA ratios and BF16 attention
intensities are:
\begin{center}
\begin{tabular}{lrr}
\toprule
Model & \(g=d/d_{\mathrm{kv}}\) &
\(I_{\mathrm{attn}}\) [FLOP/byte] \\
\midrule
EuroLLM-1.7B & \(2\) & \(2\) \\
EuroLLM-9B   & \(4\) & \(4\) \\
EuroLLM-22B  & \(6\) & \(6\) \\
\bottomrule
\end{tabular}
\end{center}
These values are far below the ridge point of either the A100 or H100.
Decode attention is therefore strongly memory-bandwidth bound even when the
projection and MLP GEMMs become compute-bound.

For the whole-model estimate, let \(N_W\) denote the number of weights that
must be streamed during one decoding step. We exclude the input embedding
table because inference reads only the embeddings of the current tokens, but
include the full output projection:
\begin{center}
\begin{tabular}{p{1.5cm}rp{1.9cm}p{1.9cm}}
\toprule
Model \newline EuroLLM
& \(N_W\)
& BF16 \newline weight bytes
& INT4 \newline weight bytes \\
\midrule
1.7B
& \(1.395\)B & \(2.79\) GB & \(0.70\) GB \\
9B
& \(8.628\)B & \(17.26\) GB & \(4.31\) GB \\
22B
& \(21.851\)B & \(43.70\) GB & \(10.93\) GB \\
\bottomrule
\end{tabular}
\end{center}
This assumes that all streamed matrices, including the output projection, are
quantized. In practice, some implementations leave embeddings or the language
model head in BF16.

In the fully memory-bound approximation, the upper-bound speedup from BF16
to INT4 weights is
\[
    \mathrm{speedup}_{\mathrm{W4}}
    \leq
    \frac{
        2N_W+Q_{\mathrm{KV}}
    }{
        0.5N_W+Q_{\mathrm{KV}}
    }.
    \label{eq:kv-limited-speedup}
\]
This approaches \(4\times\) when weight traffic dominates, but approaches
\(1\times\) when KV-cache traffic dominates. Moreover, weights can be reused
across the batch, whereas each sequence has its own KV cache. Large offline
batches therefore make KV traffic important at much shorter context lengths.

The context length at which BF16 KV traffic equals the INT4 weight traffic is
\[
    S_{\mathrm{cross}}
    =
    \frac{0.5N_W}
         {2BLd_{\mathrm{kv}}b_a}.
\]
For the concrete EuroLLM models, this gives:
\begin{center} \begin{tabular}{lrrr} \toprule Model & \(B=1\) & \(B=4\) & \(B=64\) \\ \midrule EuroLLM-1.7B & \(7{,}094\) & \(1{,}774\) & \(111\) \\ EuroLLM-9B & \(25{,}077\) & \(6{,}269\) & \(392\) \\ EuroLLM-22B & \(49{,}395\) & \(12{,}349\) & \(772\) \\ \bottomrule \end{tabular} \end{center}

For EuroLLM-1.7B, KV-cache traffic already exceeds the INT4 weight traffic
at approximately \(111\) tokens for an offline batch of 64. For EuroLLM-9B,
the corresponding crossover occurs at approximately \(392\) tokens. For
EuroLLM-22B, its larger weight matrices delay the crossover to approximately
\(772\) tokens.

At the native maximum context lengths, the memory-traffic upper bound predicts:
\begin{center} \begin{tabular}{p{3cm}rrr} \toprule Model and context & \(B=1\) & \(B=4\) & \(B=64\) \\ \midrule EuroLLM-1.7B, \newline \(S=4096\) & \(2.90\times\) & \(1.91\times\) & \(1.08\times\) \\ EuroLLM-9B, \newline \(S=32768\) & \(2.30\times\) & \(1.48\times\) & \(1.04\times\) \\ EuroLLM-22B, \newline \(S=32768\) & \(2.80\times\) & \(1.82\times\) & \(1.07\times\) \\ \bottomrule \end{tabular} \end{center}
These numbers are upper bounds based only on total HBM traffic. They already
show that a nominal \(4\times\) reduction in weight bytes need not translate
into a \(4\times\) end-to-end decode speedup.

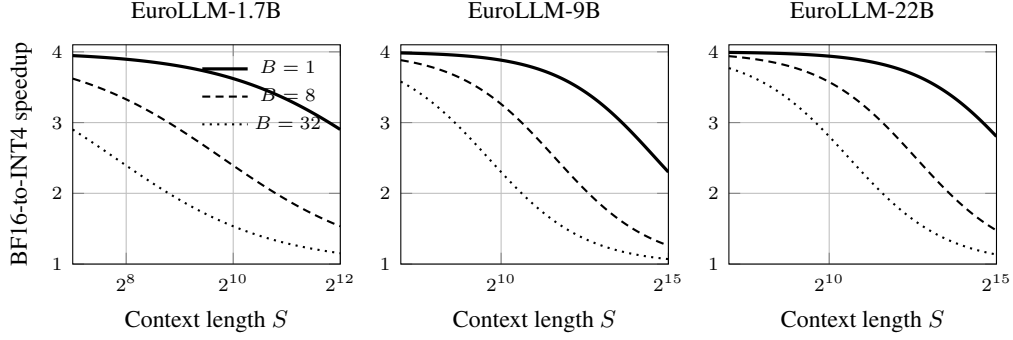
\begin{figure*}[t]
\centering
\begin{tikzpicture}
\begin{groupplot}[
    group style={
        group size=3 by 1,
        horizontal sep=0.8cm
    },
    width=0.32\textwidth,
    height=0.28\textwidth,
    xmode=log,
    log basis x=2,
    ymin=1,
    ymax=4.1,
    xlabel={Context length \(S\)},
    grid=both,
    tick label style={font=\scriptsize},
    label style={font=\small},
    title style={font=\small},
    legend style={
        font=\scriptsize,
        draw=none,
        fill=none,
        at={(0.98,0.98)},
        anchor=north east
    },
    samples=180
]

\nextgroupplot[
    title={EuroLLM-1.7B},
    xmin=128,
    xmax=4096,
    domain=128:4096,
    ylabel={BF16-to-INT4 speedup}
]

\addplot[black, very thick]
{
    (2*1.394706432e9 + 4*1*24*x*1024)
    /
    (0.5*1.394706432e9 + 4*1*24*x*1024)
};
\addlegendentry{\(B=1\)}

\addplot[black, thick, densely dashed]
{
    (2*1.394706432e9 + 4*8*24*x*1024)
    /
    (0.5*1.394706432e9 + 4*8*24*x*1024)
};
\addlegendentry{\(B=8\)}

\addplot[black, thick, dotted]
{
    (2*1.394706432e9 + 4*32*24*x*1024)
    /
    (0.5*1.394706432e9 + 4*32*24*x*1024)
};
\addlegendentry{\(B=32\)}

\nextgroupplot[
    title={EuroLLM-9B},
    xmin=128,
    xmax=32768,
    domain=128:32768
]

\addplot[black, very thick]
{
    (2*8.628031488e9 + 4*1*42*x*1024)
    /
    (0.5*8.628031488e9 + 4*1*42*x*1024)
};

\addplot[black, thick, densely dashed]
{
    (2*8.628031488e9 + 4*8*42*x*1024)
    /
    (0.5*8.628031488e9 + 4*8*42*x*1024)
};

\addplot[black, thick, dotted]
{
    (2*8.628031488e9 + 4*32*42*x*1024)
    /
    (0.5*8.628031488e9 + 4*32*42*x*1024)
};

\nextgroupplot[
    title={EuroLLM-22B},
    xmin=128,
    xmax=32768,
    domain=128:32768
]

\addplot[black, very thick]
{
    (2*21.850896384e9 + 4*1*54*x*1024)
    /
    (0.5*21.850896384e9 + 4*1*54*x*1024)
};

\addplot[black, thick, densely dashed]
{
    (2*21.850896384e9 + 4*8*54*x*1024)
    /
    (0.5*21.850896384e9 + 4*8*54*x*1024)
};

\addplot[black, thick, dotted]
{
    (2*21.850896384e9 + 4*32*54*x*1024)
    /
    (0.5*21.850896384e9 + 4*32*54*x*1024)
};

\end{groupplot}
\end{tikzpicture}

\caption{
Memory-bandwidth upper bound on the speedup from BF16 to INT4 weights for
the three EuroLLM models with a BF16 KV cache. Increasing context length or
batch size increases KV-cache traffic, which is unaffected by weight-only
quantization. The relative memory-bound speedup is independent of the GPU's
absolute HBM bandwidth, although absolute latency is lower on the H100.
}
\label{fig:eurollm-kv-limited-weight-quantization}
\end{figure*}

This explains why weight-only quantization does not always yield the speedup
suggested by the reduction in model size. At small decode batches and short
contexts, weight traffic dominates and the GEMM-only roofline predicts a large
gain. At larger batches, weights are amortized across tokens and the GEMMs may
become compute-bound. At long contexts, KV-cache reads and attention dominate
the remaining memory traffic, so reducing weight bytes has progressively less
effect.

Finally, the idealized model ignores the cost of unpacking and dequantizing
weights, quantization scales and zero points, small-\(M\) GEMM underutilization,
kernel-launch overhead, normalization and positional-encoding kernels,
paged-KV-cache indirections, communication, and sampling. It also assumes
that every matrix is quantized and that all HBM traffic reaches peak bandwidth.
These effects can make the realized speedup substantially smaller than the
roofline upper bound.

\subsection{Details on Quantization Formats}\label{appendix:details_quantization}

We provide additional details to the three quantization formats that we consider in this work: W8A8, W4A8, and W4A16 weight quantization. 
In W8A8, the weights are rescaled by the absmax of each row along the output dimension, while the activations are similarly quantized dynamically along the token dimension. As discussed in \cite{xiao2023smoothquant}, this maps well to hardware-accelerated GEMM kernels, as one can apply the scaling factors after the matrix multiplication:
\begin{equation*}
    \mathbf{Y} = \text{diag}(\Delta_{\mathbf{X}}^{\text{BF16}}) \cdot (\bar{\mathbf{X}}^{\text{INT8}} \cdot \bar{\mathbf{W}}^{\text{INT8}}) \cdot \text{diag}(\Delta_{\mathbf{W}}^{\text{BF16}}),
\end{equation*}
where $\mathbf{Y} \in \R^{d_\text{in} \times d_\text{out}}$ refers to the output matrix, $\bar{\mathbf{X}}^{\text{INT8}} \in \R^{d_\text{in} \times n}$ and $\bar{\mathbf{W}}^{\text{INT8}} \in \R^{n \times d_{\text{out}}}$ refer to the quantized activation and weight matrices, stored in INT8, and $\Delta_{\mathbf{X}}^{\text{BF16}} \in \R^{d_\text{in}}$ and $\Delta_{\mathbf{W}}^{\text{BF16}} \in \R^{d_\text{out}}$ refer to the activation scale vector, respectively the weight scale vector, stored in BFloat16.

For W4A8, weights are quantized with a group size of 128 following \cite{zhang2024qqq}, while activations are dynamically quantized along the token dimension.
Finally for W4A16 weight quantization, weights are also quantized with a group size of 128 while activations remain unchanged.
All linear layers in the model, except for the embedding and prediction head layer, are being quantized using the same data format. 

\subsection{Details on document-level evaluation}\label{appendix:details_chunklen_eval}

We provide additional details on the document-level evaluation in this section. 
The prompt template used for the evaluation of both models can be found in \cref{tab:prompt_templates}, which we took from \cite{martins2025eurollm}.
We choose a temperature of $T=0$, $\text{top}_p = 0.8$, and $\text{max\_tokens} = 4096$ as decoding parameters and evaluate document chunking for target prompt-length thresholds in $\{100, 400, 800, 1000, 1500\}$ tokens for EuroLLM-1.7B and in $\{100, 400, 800, 1000, 2000, 4000\}$ tokens for EuroLLM-9B on DocHPLT. On WMT24++ we evaluate document chunking for target prompt-length thresholds in $\{100, 200, 400, 500, 600, 800, 1000\}$. 
We note that we do not split segments and keep trailing chunks. This means that the chunk lengths can overshoot on some chunks and documents can end with a chunk much shorter than the threshold.
Note that the target thresholds also include the rendered chat template. 
Across the four translation directions we have 27 matched documents for UK $\to$ EN, 13 documents for EN $\to$ UK from DocHPLT, as well as 19 matched documents for DE $\to$ EN and 11 documents for EN $\to$ DE from WMT24++. The common document subset was formed by performing an intersection of valid documents across all models, formats and target thresholds for each translation direction.
We compute the chrF++ scores by setting $\text{word\_order}=2$. The transformers version used for the experiments was 5.14.1

\begin{table*}[t]
\centering
\caption{Prompt templates for evaluating EuroLLM.}
\label{tab:prompt_templates}

\vspace{0.5em}

\paragraph{EuroLLM-Instruct} \mbox{}\\
\begin{tabularx}{\textwidth}{l X}
\toprule
\textbf{Role} & \textbf{Prompt Template} \\
\midrule
System & Translate all user texts to \texttt{\{target\_lang\}}. \\
User & \texttt{\{source\_text\}} \\
\bottomrule
\end{tabularx}
\end{table*}

\subsection{Further Results on Chunk Length vs. Translation Quality}\label{appendix:results_chunklen_chrF}

\begin{figure*}[ht!]
    \centering
    \includegraphics[width=0.7\linewidth]{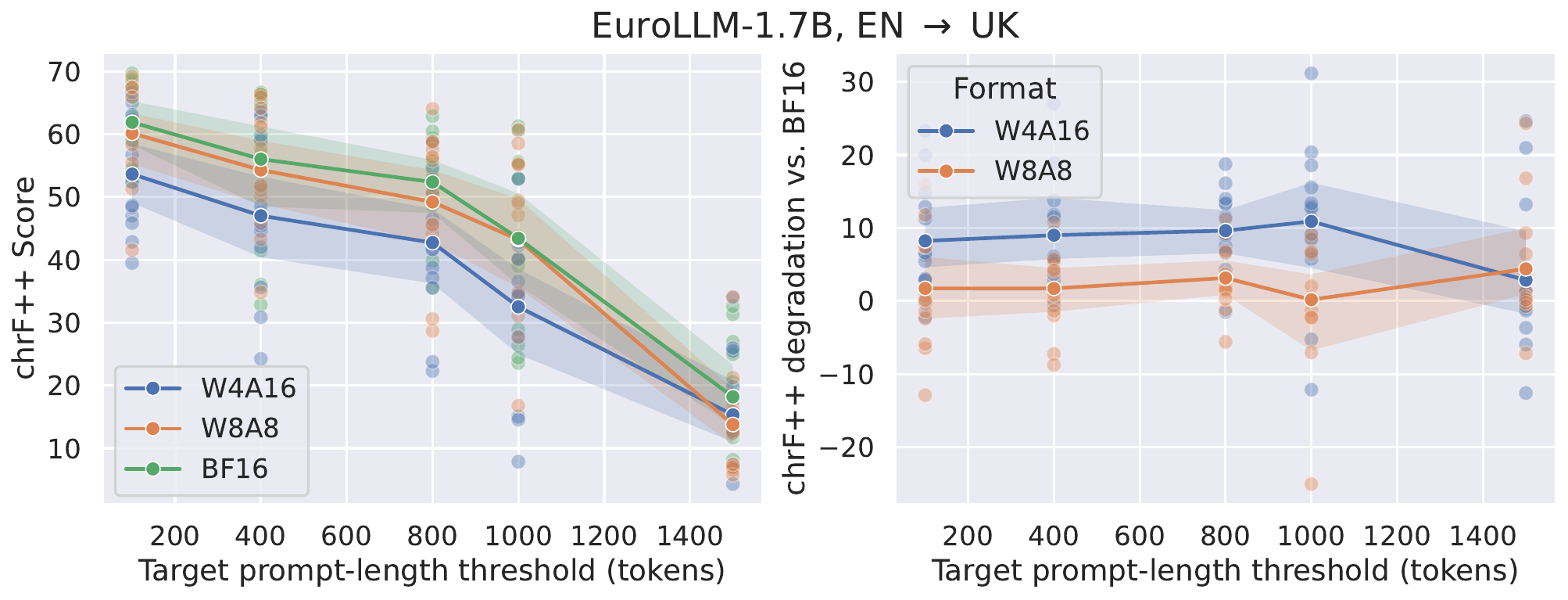}
    \includegraphics[width=0.7\linewidth]{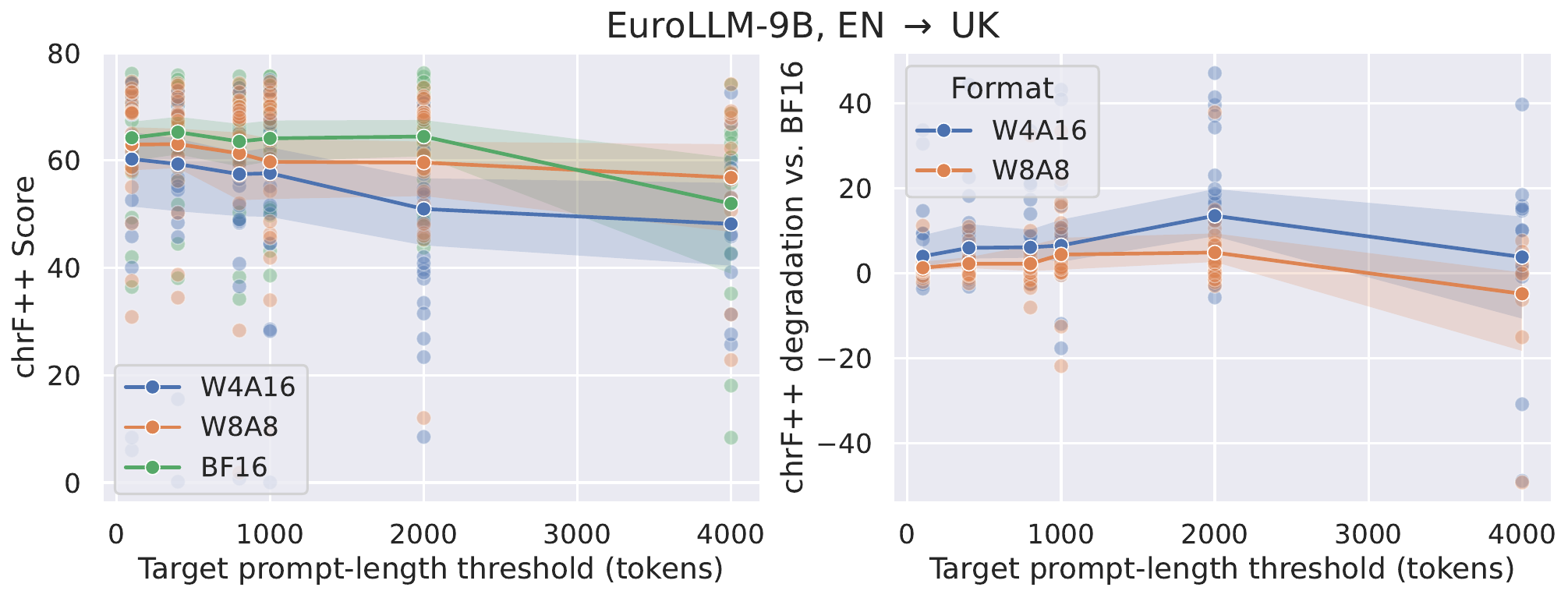}
    \caption{Chunk length vs. chrF++ Score for EuroLLM-1.7B (top row) and EuroLLM-9B (bottom row) across different quantization formats averaged across a subset of 13 documents from DocHPLT. Shaded area corresponds to bootstrapped 95\% CI of document-level means. Each point denotes a document-level chrF++ score in the left subplots and a paired BF16-quantized difference in the right subplots. Note the different $y$ limits for better visibility.}
    \label{fig:chunklen_overview_EN_UK}
\end{figure*}

\begin{figure*}[ht!]
    \centering
    \includegraphics[width=0.7\linewidth]{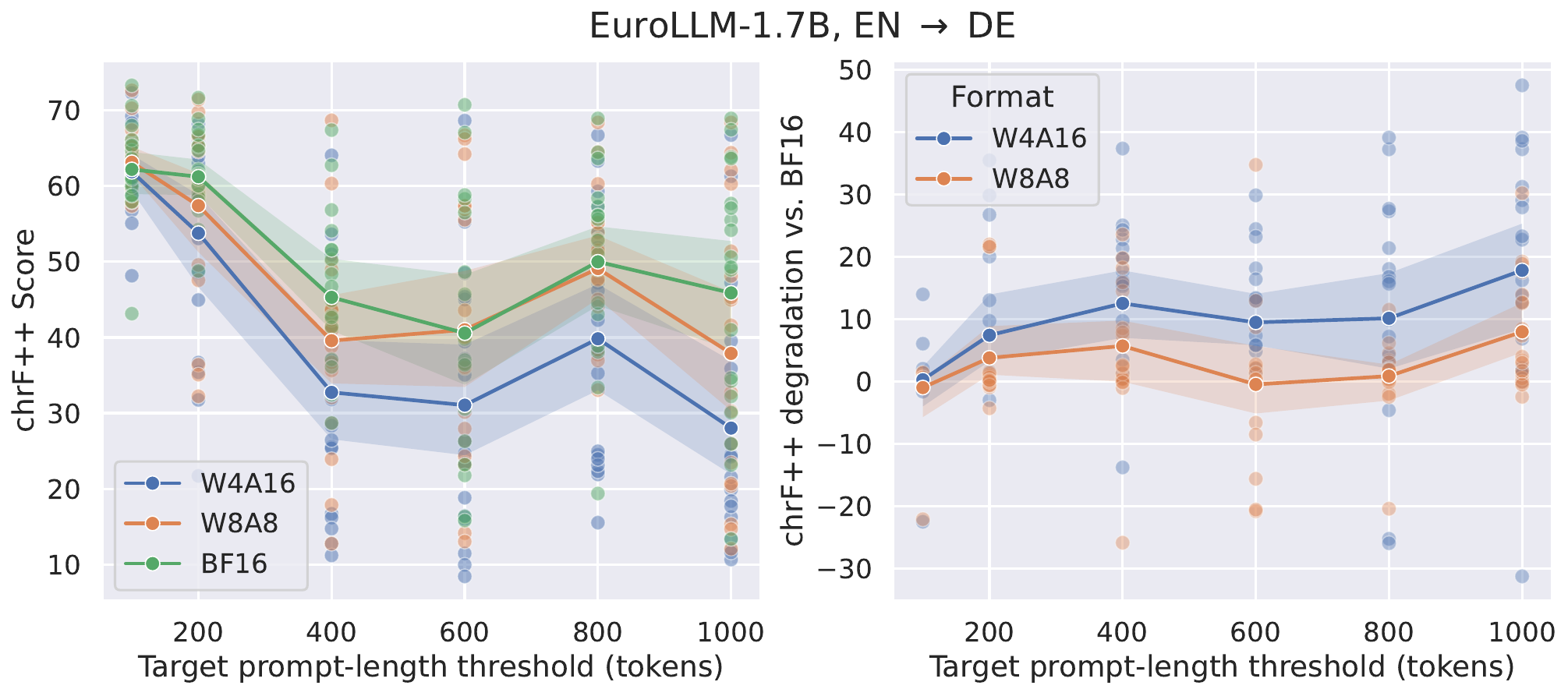}
    \includegraphics[width=0.7\linewidth]{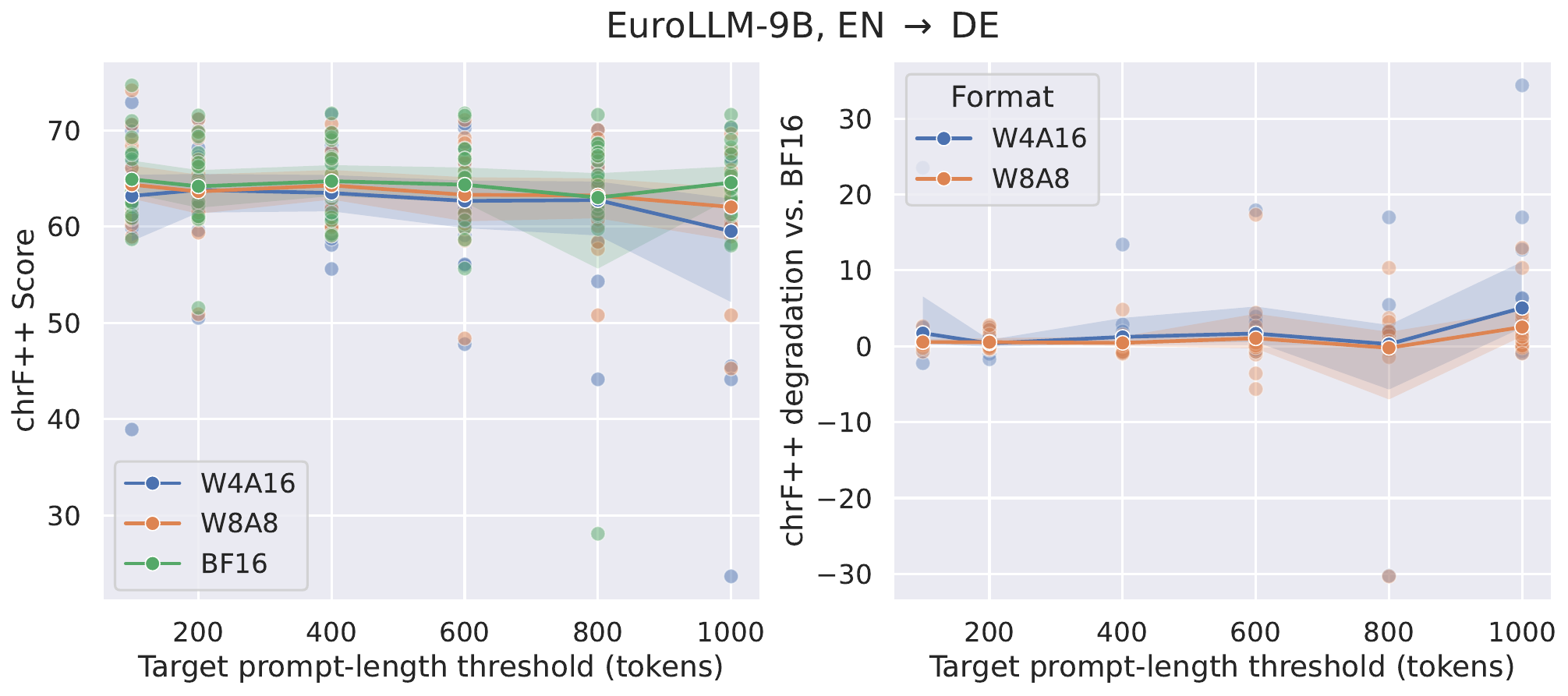}
    \caption{Chunk length vs. chrF++ Score for EuroLLM-1.7B (top row) and EuroLLM-9B (bottom row) across different quantization formats averaged across a subset of 11 documents from WMT24++. Shaded area corresponds to bootstrapped 95\% CI of document-level means. Each point denotes a document-level chrF++ score in the left subplots and a paired BF16-quantized difference in the right subplots. Note the different $y$ limits for better visibility.}
    \label{fig:chunklen_overview_EN_DE}
\end{figure*}

\begin{figure*}[ht!]
    \centering
    \includegraphics[width=0.7\linewidth]{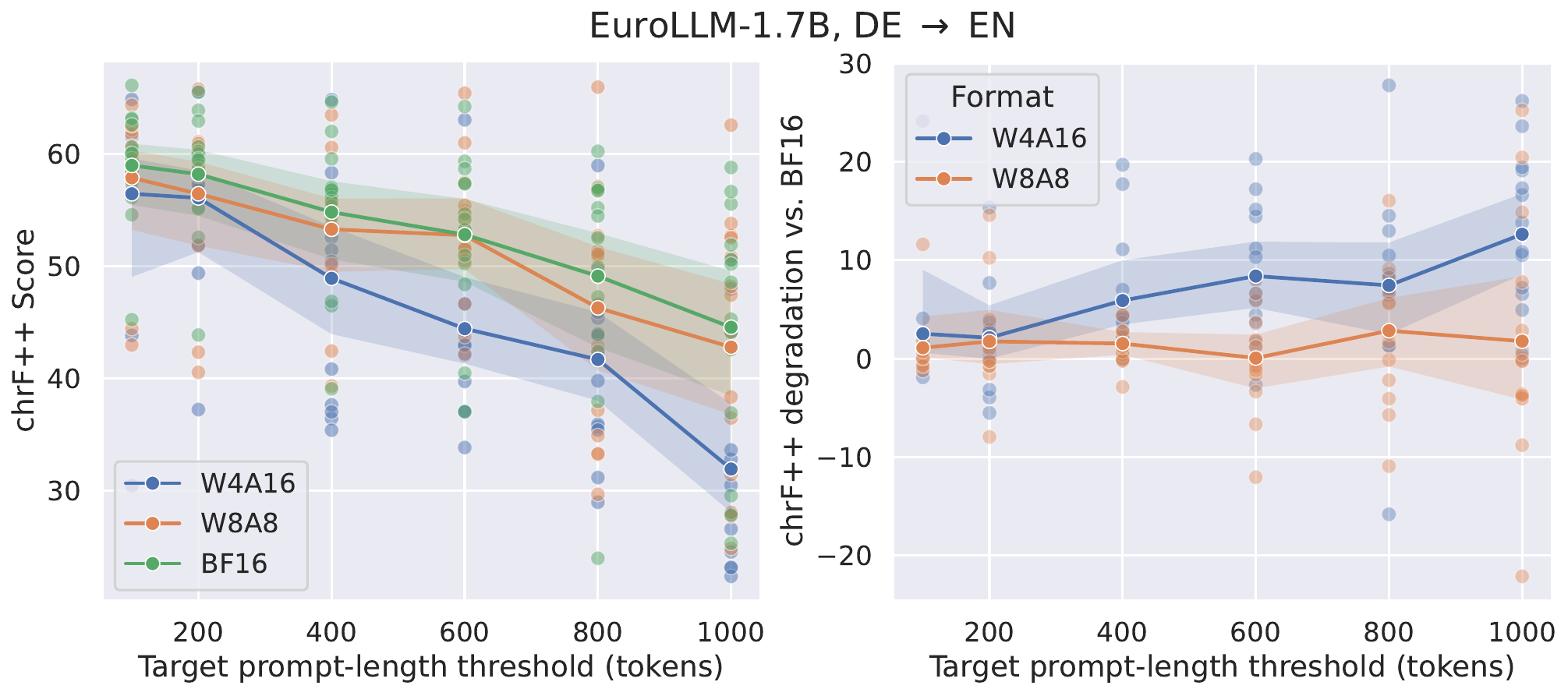}
    \includegraphics[width=0.7\linewidth]{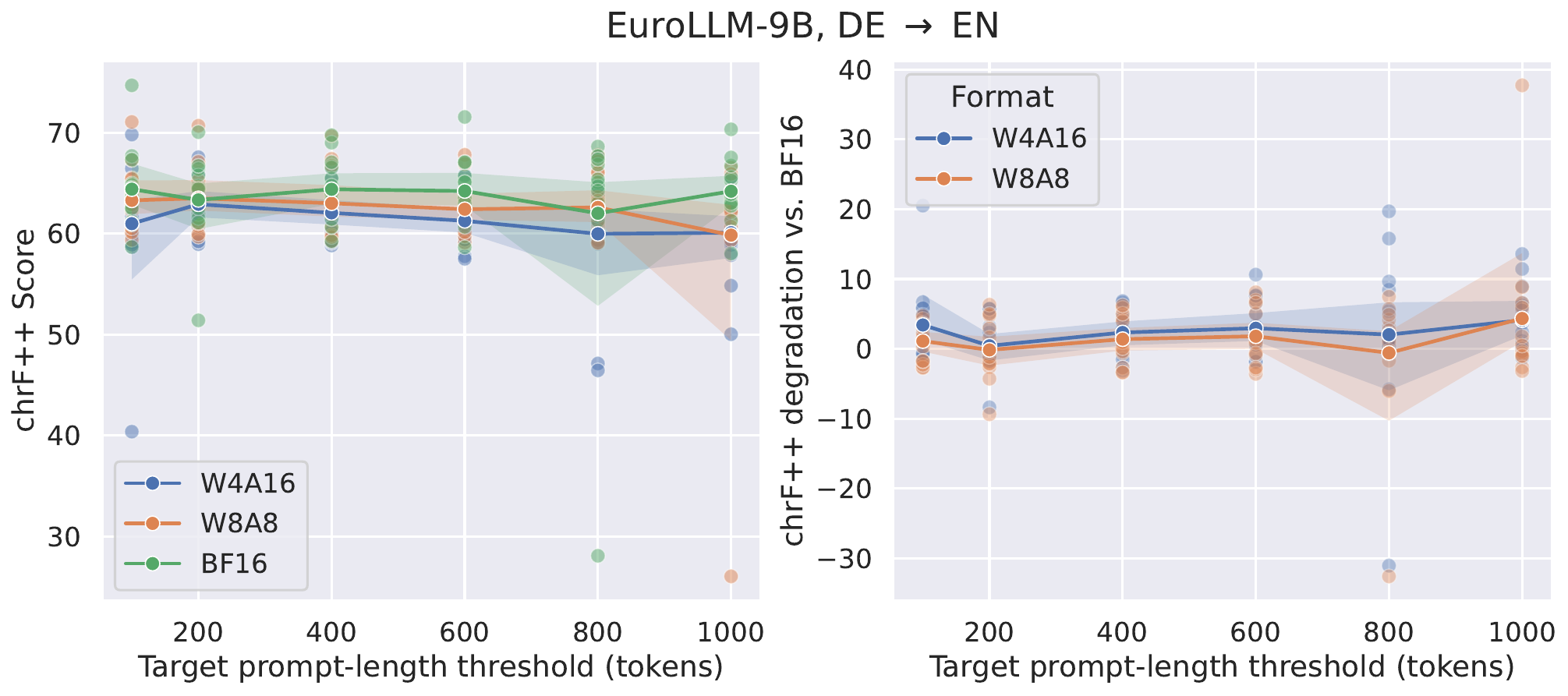}
    \caption{Chunk length vs. chrF++ Score for EuroLLM-1.7B (top row) and EuroLLM-9B (bottom row) across different quantization formats averaged across a subset of 19 documents from WMT24++. Shaded area corresponds to bootstrapped 95\% CI of document-level means. Each point denotes a document-level chrF++ score in the left subplots and a paired BF16-quantized difference in the right subplots. Note the different $y$ limits for better visibility.}
    \label{fig:chunklen_overview_DE_EN}
\end{figure*}

We provide further experimental results on how chunk length affects translation quality for English to Ukrainian translation on a subset of DocHPLT (\cref{fig:chunklen_overview_EN_UK}). Earlier experiments on EN $\leftrightarrow$ DE on WMT24++ can be found in \cref{fig:chunklen_overview_EN_DE} and \cref{fig:chunklen_overview_DE_EN}. The results show that whether an increased degradation with context length depends not only on the quantization format and translation direction, but also on the model itself. Note that WMT24++ contains considerably shorter documents, such that translation quality is only evaluated up to a threshold of 1000 tokens limiting its power to say anything for long document translations.

\subsection{Further Results on Offline Benchmarking}\label{appendix:results_offline_benchmarking}

\begin{figure*}[ht!]
    \centering
    \includegraphics[width=0.9\linewidth]{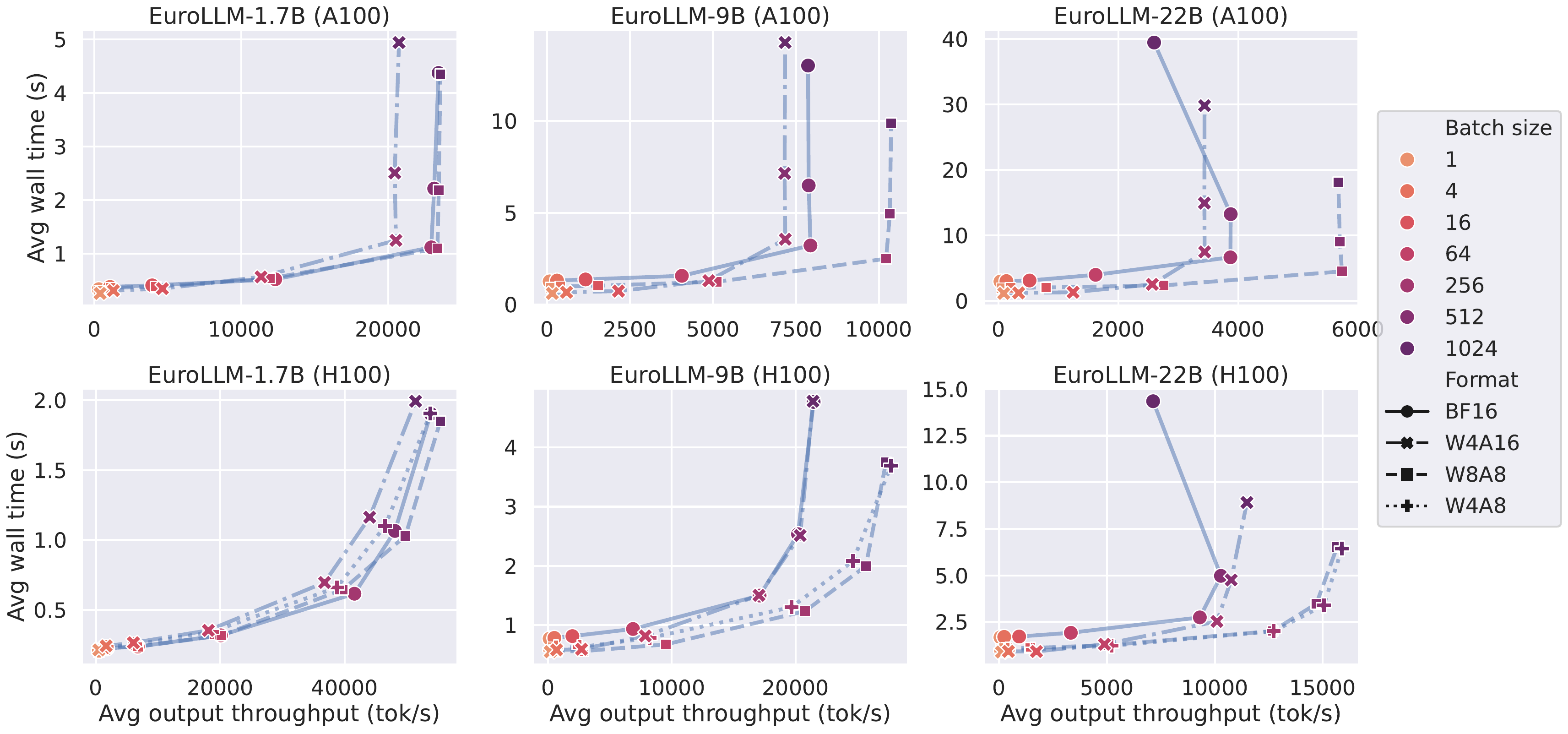}
    \caption{Offline benchmark measuring average output throughput in tok/s against average wall time for a sequence length of \textit{100 tokens} across three model sizes on a A100 ({top row}) and a H100 ({bottom row}). }
    \label{fig:offline_benchmark_pareto_curve_overview_100}
\end{figure*}

We provide further results on offline benchmarking models from the EuroLLM family for sequence lengths of 100 tokens in \cref{fig:offline_benchmark_pareto_curve_overview_100}.  

\end{document}